%% file: main.tex
\documentclass[runningheads]{llncs}

\usepackage{eccv}

\usepackage{eccvabbrv}

\usepackage{graphicx}
\usepackage{booktabs}
\usepackage{subcaption}
\usepackage{array}
\usepackage{wrapfig}

\usepackage[accsupp]{axessibility}  

\usepackage{hyperref}

\usepackage{orcidlink}
\usepackage{multirow}
\usepackage{graphicx}
\usepackage{subcaption}
\usepackage{booktabs} 
\usepackage{tikz}
\usetikzlibrary{arrows.meta,positioning,fit,calc}
\usepackage{graphicx,xcolor}
\usepackage{tikz}
\usepackage{amssymb}    
\usepackage[most]{tcolorbox}

\usepackage[table]{xcolor}
\usepackage{bm}
\usepackage{subcaption}

\definecolor{gtgreen}{RGB}{93, 207, 95}
\definecolor{wrongred}{RGB}{231,76,60}
\definecolor{editorange}{RGB}{243,156,18}
\definecolor{refblue}{RGB}{52,152,219}

\begin{document}

\title{Error‑Driven Scene Editing for 3D Grounding \\in Large Language Models} 

\titlerunning{DEER-3D}

\author{Yue Zhang\inst{1}\orcidlink{0000-0003-2153-6536} \and
Zun Wang\inst{1}\orcidlink{0009-0005-9502-050X} \and
Han Lin\inst{1}\orcidlink{0000-0002-1303-8636}  \and
Jialu Li\inst{1}\orcidlink{0009-0001-8807-2644
} \and
Jianing Yang\inst{2}\orcidlink{0000-0003-0297-7889} \and \\ 
Yonatan Bitton\inst{3} \and
Idan Szpektor\inst{3} \and
Mohit Bansal\inst{1}\orcidlink{0009-0009-2965-5354}}

\authorrunning{Y. Zhang et al.}

\institute{University of North Carolina at Chapel Hill, USA  \and
University of Michigan, USA \and
Google Research, USA\\
\url{https://github.com/zhangyuejoslin/Deer-3D}
}

\maketitle

\input{sec/0_abstract}    
\input{sec/1_intro}

\input{sec/2_related_work}
\input{sec/3_method}

\input{sec/4_experiment}

\input{sec/5_conclusion}

%
%
\bibliographystyle{splncs04}
\bibliography{main}

\clearpage
\input{sec/X_suppl}

\end{document}

%% file: sec/0_abstract.tex
\begin{abstract}
Despite recent progress in 3D-LLMs, they remain limited in accurately grounding language to visual and spatial elements in 3D environments. This limitation stems in part from training data that focuses on language reasoning rather than spatial understanding due to scarce 3D resources, leaving inherent grounding biases unresolved.
To address this, we propose 3D scene editing as a key mechanism to generate visual counterfactuals that mitigate these biases through fine-grained spatial manipulation, without requiring costly scene reconstruction or large-scale 3D data collection. 
Furthermore, to make these edits targeted and directly address the specific weaknesses of the model, we introduce \textbf{DEER-3D}, an error-driven framework that diagnoses grounding failures and generates targeted counterfactual training supervision via a structured ``Decompose, Diagnose, Edit, and Retrain” loop. 
Specifically, given a grounding failure, DEER-3D first identifies the predicate-level error (e.g., attribute or spatial relation). 
It then performs minimal predicate-aligned scene edits, such as recoloring or repositioning, and constructs aligned question–answer pairs that explicitly target the failed predicate, forming targeted counterfactual training examples.
We evaluate our editing pipeline across multiple benchmarks for 3D grounding and scene understanding tasks, consistently demonstrating improvements across all grounding datasets through iterative refinement (4-6\% gains). DEER-3D underscores the effectiveness of targeted, error-driven scene editing in bridging linguistic reasoning with spatial grounding in 3D LLMs.
\keywords{3D Visual Grounding \and 3D Large Language Models \and Error-Driven Framework}
\end{abstract}


%% file: sec/1_intro.tex
\section{Introduction}
\label{sec:intro}

Grounding language in 3D environments, identifying the specific objects, attributes, and spatial relations that a description refers to, is fundamental for embodied AI and robot manipulation~\cite{ma2024llms,zhang2024vision, hu20253dllm, cheng2024spatialrgpt}.
Recent efforts have integrated large language models (LLMs) with 3D perception, giving rise to so-called \textit{3D-LLMs}~\cite{wang2023chat, hong20233d,huang2023embodied,huang2024chat,linghu2024multi,zhangspartun3d,deng20253d, xu2024pointllm}. 
These models have demonstrated impressive progress in open-vocabulary captioning, question answering, and 3D scene reasoning, largely benefiting from the strong linguistic capabilities of LLMs. 
However, despite these achievements, the grounding performance of current 3D-LLMs remains limited, failing to localize fine-grained visual details (e.g., confusing two pillows of different colors), misinterpreting spatial relations (e.g., confusing ``far from'' as ``near''), or rely on language priors rather than geometric evidence (e.g., grounding ``pillow'' on the ``bed'' even when the instruction describes a pillow on the floor)~\cite{huang2025unveiling, zhang2025point}. 
This gap highlights that linguistic proficiency alone does not guarantee spatially grounded understanding. Fundamentally, it reveals a critical cross-modality alignment challenge: mapping discrete textual semantics (e.g., “the chair facing the window'') onto the continuous geometric representations of 3D point clouds composed of tens of thousands of points.
 
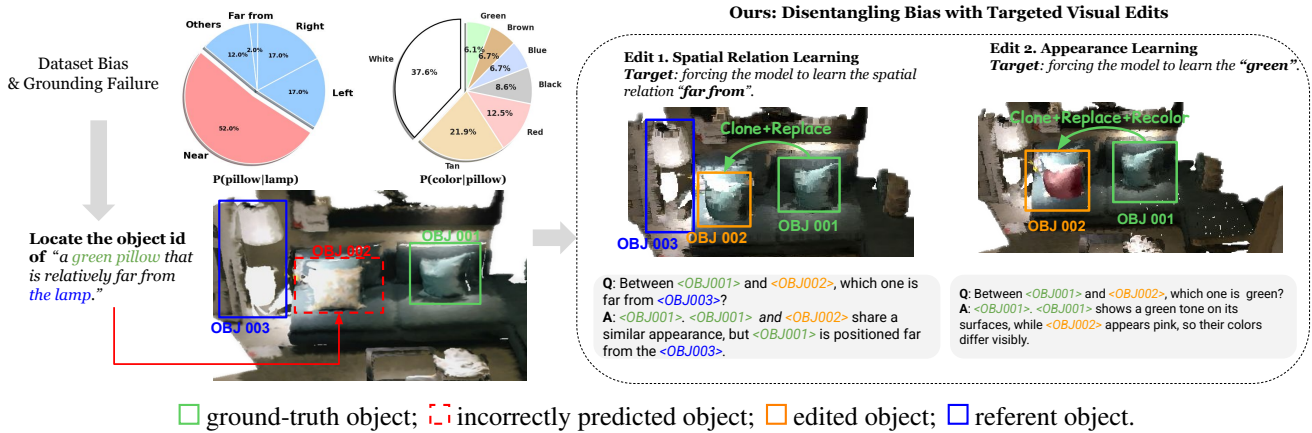
\begin{figure*}[t]
    \centering
\includegraphics[width=\linewidth]{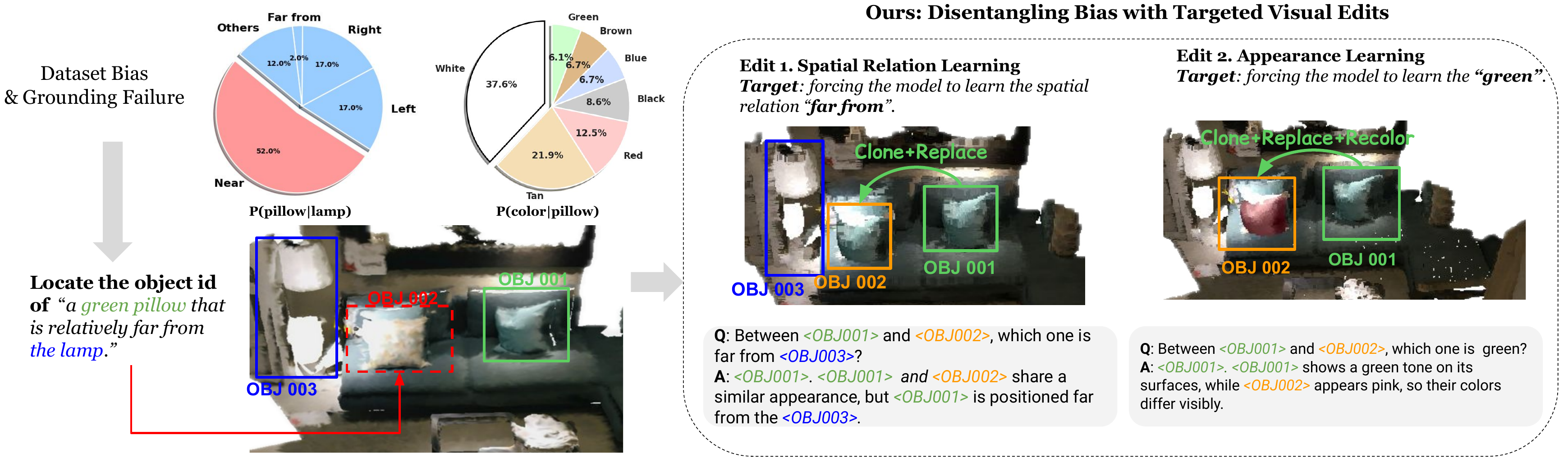}
\tikz{\draw[line width=0.8pt, color=gtgreen] (0,0) rectangle (0.25,0.25);} target-obj;\,
    \tikz{\draw[dashed, line width=0.8pt, color=red] (0,0) rectangle (0.25,0.25);} incorrectly-predicted-obj;\, 
    \tikz{\draw[line width=0.8pt, color=orange] (0,0) rectangle (0.25,0.25);} edited-obj;\, 
    \tikz{\draw[line width=0.8pt, color=blue] (0,0) rectangle (0.25,0.25);} referent-obj.
    \caption{3D-LLMs frequently overfit to dataset co-occurrence biases (e.g., white pillows), causing grounding failures.
We mitigate these biases through targeted counterfactual edits in 3D scenes and construct aligned QA pairs to strengthen the model’s grounding.}
    \label{fig:teaser}
\end{figure*}

Understanding why current 3D-LLMs struggle with grounding requires examining both dataset biases and how supervision signals are structured during training. Due to the scarcity of large-scale 3D resources, existing datasets often contain latent statistical biases that models exploit as shortcuts~\cite{zhang2025point}. ~\cref{fig:teaser} illustrates an example that over half of the spatial relationships between ``lamp'' and ``pillow'' are categorized as ``near,'' and most pillows are labeled as ``white.'' Consequently, models learn superficial associations (e.g., ``white pillows are near lamps'') rather than genuine geometric relationships, leading to failures such as incorrectly grounding ``a green pillow far from the lamp.''
However, current approaches for improving 3D-LLMs primarily focus on text-based augmentations~\cite{huang2023embodied, huang20253d, zhangspartun3d, linghu2024multi}, which inherently cannot resolve such visual biases. While such strategies improve general linguistic understanding, they leave the underlying 3D scenes unchanged and fail to introduce the necessary visual variations to challenge these spurious co-occurrence patterns, and can even exacerbate them by reinforcing the statistical priors.
Moreover, conventional augmentations are typically applied ``blindly,”  which simultaneously modifies several predicates (e.g., color or spatial relation), mixing supervision signals across factors and making it harder for the model to learn which factor actually matters.

To address these limitations, we propose \textbf{DEER-3D} (\textbf{D}ecompose, Diagnostic \textbf{E}valuation, \textbf{E}dit, and \textbf{R}etrain),  a structured framework for improving 3D grounding through error diagnosis and targeted scene editing.
DEER-3D first attributes grounding failures to specific semantic predicates under a deterministic evaluation protocol. It then constructs controlled 3D counterfactual scenes that modify the failed predicate while minimizing changes to other geometric factors. 
The edited scenes are paired with predicate-aligned question–answer pairs that explicitly probe the modified semantics, forming targeted counterfactual training examples. 
This predicate-level visual intervention, together with aligned QA, provides clearer supervision signals, encouraging the model to rely on genuine geometric evidence instead of dataset co-occurrence biases.

Specifically, to mitigate the semantic-factor entanglement underlying grounding failures, DEER-3D systematically performs three structured stages (see~\cref{fig:main_fig}).
In the \textbf{Decompose stage}(\cref{fig:main_fig}(a)), \textsc{DEER}-3D breaks down the complex natural-language query into atomic sub-instructions, which are then deterministically mapped to a predefined set of semantic categories. In the \textbf{Diagnose stage}~(~\cref{fig:main_fig}(b)), it systematically queries the base 3D-LLM with each isolated sub-instruction. 
If the model fails under a specific atomic predicate, \textsc{DEER}-3D explicitly identifies the grounding error to that semantic factor (e.g., color, orientation, or distance), enabling targeted counterfactual intervention.
In the subsequent \textbf{Edit stage}(\cref{fig:main_fig}(c)), \textsc{DEER}-3D performs a duplicate-and-replace operation: it duplicates the target object geometry and substitutes a nearby distractor with this duplicate, while minimizing unintended geometric alterations so that the modification primarily affects the failed semantic predicate. A minimal visual modification, such as recoloring or rotating the duplicate, creates a counterfactual pair that explicitly isolates the erroneous appearance or spatial relation, providing fine-grained supervision signals. We further pair these edited scenes with corresponding question-answer pairs that capture different levels of reasoning difficulty.
Finally, in the \textbf{Retrain stage}(\cref{fig:main_fig}(d)), these targeted counterfactual examples are integrated back into the training set. Through iterative refinement, the model is exposed to progressively constructed predicate-targeted counterfactual supervision signals, improving robustness and predicate-specific grounding accuracy.

We extensively evaluate \textsc{DEER}-3D on established 3D visual grounding benchmarks and observe substantial improvements across all metrics (e.g., achieving 4-6\% gains in grounding accuracy), clearly demonstrating its effectiveness in resolving fine-grained grounding errors. Crucially, we validate the benefit of our iterative refinement framework by conducting multi-round experiments, illustrating how progressively targeting model failures further boosts performance. 
Comprehensive ablation studies confirm the effectiveness of our targeted visual editing strategies. In particular, our error-driven counterfactual supervision outperforms random augmentation even under fewer edited samples, and a text-only diagnostic variant yields only marginal gains, confirming that predicate-level visual intervention is essential for resolving visually entangled grounding failures.
Additionally, we evaluate \textsc{DEER}-3D's broader utility beyond grounding, demonstrating consistent improvements on general 3D tasks, as well as on human-aligned grounding evaluations. Taken together, these experiments underscore the effectiveness and robustness of our approach in enhancing the grounding capabilities of 3D-LLMs.

%% file: sec/2_related_work.tex
\section{Related Work}
\label{sec:related work}

\noindent\textbf{3D-LLMs for Grounding.} 
Recent advances in 3D-LLMs have greatly expanded the paradigm of 3D grounding by coupling LLMs with 3D visual perceptions, enabling open-vocabulary understanding and spatial reasoning in 3D scene environments~\cite{hong20233d, huang2023embodied, huang20253d, zhangspartun3d, wang2023chat, zhi2025lscenellm, chen2024grounded, zhu2024unifying, deng20253d, yu2025inst3d,li2025seeground, yuan2024visual}. Despite their impressive reasoning ability, current 3D-LLMs often suffer from visual bias, relying on linguistic priors rather than genuine 3D evidence when making predictions.
A primary reason for this vulnerability is the community's reliance mainly on text-based augmentation to compensate for data scarcity~\cite{huang2023embodied, zhangspartun3d, linghu2024multi,hu20263dllm}, improving linguistic diversity but leaving the underlying 3D visual data unchanged. This not only fails to correct existing dataset biases but can inadvertently exacerbate them by reinforcing the model's reliance on spurious linguistic correlations.   
In contrast, we address this limitation from the visual side through an error‑driven visual augmentation framework that explicitly edits 3D scenes to generate counterfactual pairs targeting the exact predicates responsible for grounding failures, providing fine‑grained supervision signals to correct visual bias.

\noindent\textbf{Counterfactual Augmentation in Vision and Language.}
Counterfactual data augmentation has been widely adopted to enhance robustness and causal reasoning in NLP~\cite{zmigrod2019counterfactual,maudslay2019s,kaushik2019learning,gardner2020evaluating}, vision~\cite{sauer2021counterfactual,uwaeze2025generative, melistas2024benchmarking}, robotics~\cite{parvaneh2020counterfactual,fu2020counterfactual,wang2022counterfactual, Li2022EnveditEE}, and multimodal tasks~\cite{chen2020counterfactual,teney2020learning,lai2024improving}. 
However, existing approaches are largely confined to 2D or textual domains. Extending counterfactual generation to 3D is non-trivial due to the cross-modality alignment gap between discrete linguistic predicates and continuous geometric representations. Prior methods are not designed to perform targeted, predicate-level interventions on complex 3D spatial properties such as orientation or continuous distance.

\noindent\textbf{Error-Driven Learning.} Error-driven and self-corrective learning has been applied in NLP~\cite{an2023learning, wang2023learning, yang2023failures, Zala2024EnvGenGA, Khan2024DataEnvGymDG}, multimodal models~\cite{yao2025error, wang2024targeted}, and robotics~\cite{duan2024aha},  where models iteratively improve using textual feedback or retraining on failure cases. 
In contrast, we unify error-driven refinement with counterfactual 3D scene editing, forming an iterative visual-domain loop that identifies grounding failures and applies minimal spatial interventions to progressively enhance 3D grounding in large language models.

\noindent\textbf{3D Scene Editing.} Although recent 3D scene editing methods~\cite{yan20243dsceneeditor, lee2025editsplat, wen2025intergsedit, haque2023instruct, song2023sc, zhuang2023dreameditor} enable high-fidelity and interactive object manipulation, their design goals differ fundamentally from the requirements of our counterfactual training pipeline. These systems prioritize photorealistic rendering and user-driven editing, rather than producing semantically controlled modifications that preserve all other scene attributes. In contrast, \textsc{DEER}-3D requires predicate-targeted interventions, while keeping the rest of the scene unchanged. 

%% file: sec/3_method.tex
\section{Method}

\begin{figure*}[t]
    \centering
    \includegraphics[width=\linewidth]{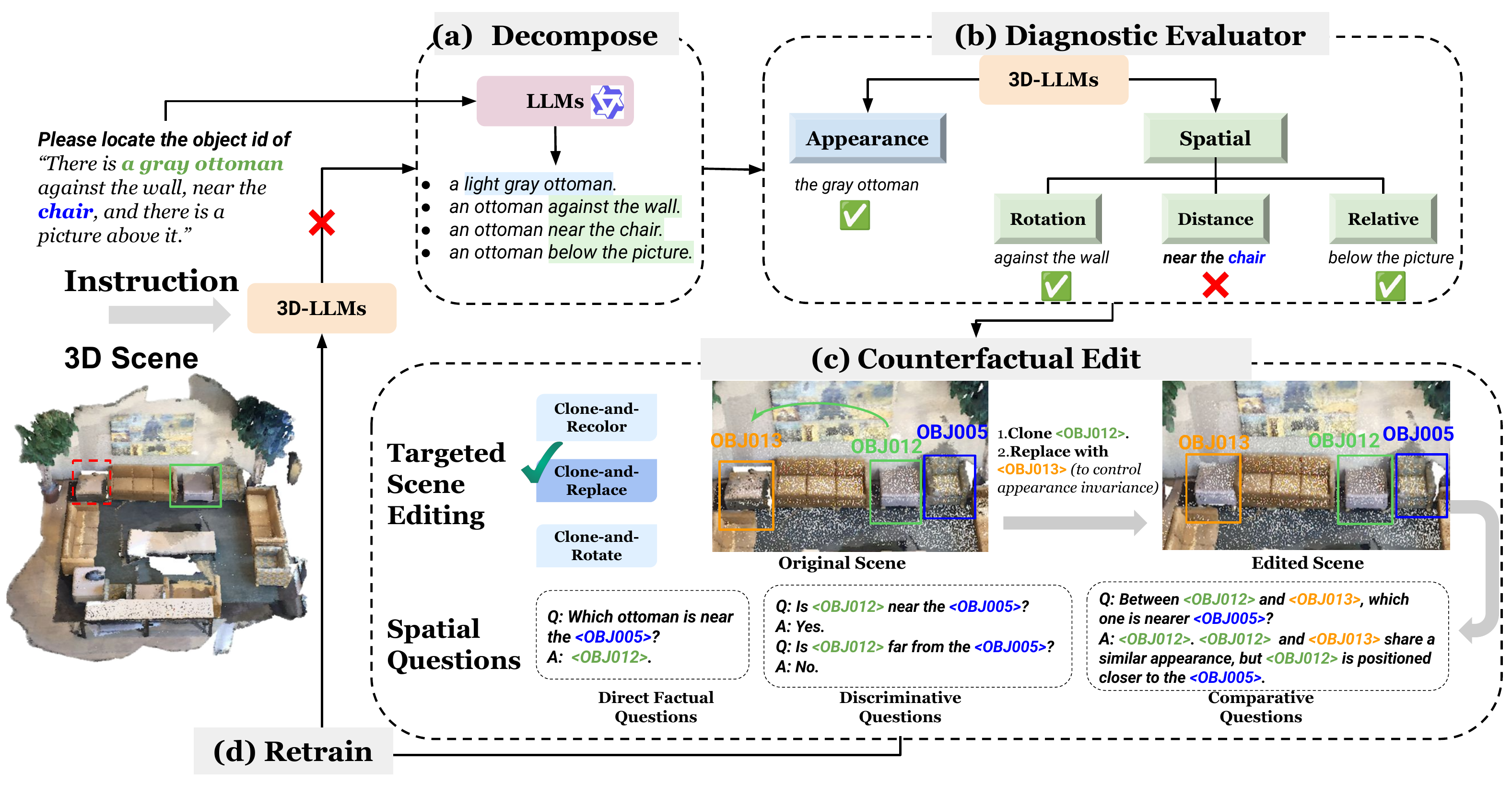}
    \small
    \tikz{\draw[line width=0.8pt, color=gtgreen] (0,0) rectangle (0.25,0.25);} targeted;\,
    \tikz{\draw[dashed, line width=0.8pt, color=red] (0,0) rectangle (0.25,0.25);} incorrectly-predicted;\,
    \tikz{\draw[line width=0.8pt, color=orange] (0,0) rectangle (0.25,0.25);} edited;\,
    \tikz{\draw[line width=0.8pt, color=blue] (0,0) rectangle (0.25,0.25);} referent
    \tikz{
    \draw[line width=0.8pt, color=red] (0,0) -- (0.25,0.25);
    \draw[line width=0.8pt, color=red] (0.25,0) -- (0,0.25);
} error detected
    \caption{
    Overview of \textsc{DEER-3D}. When a grounding error is detected, \textsc{DEER-3D} performs targeted visual edits.
    Given a natural-language instruction, the framework (a) decomposes it into atomic predicates, (b) diagnoses the specific error, and (c) applies predicate-level visual edits. Aligned question–answer pairs are then created to explicitly supervise the failed predicate. Finally, the model (d) iteratively retrains on these counterfactual examples to progressively improve grounding accuracy.
    }
    \label{fig:main_fig}
\end{figure*}
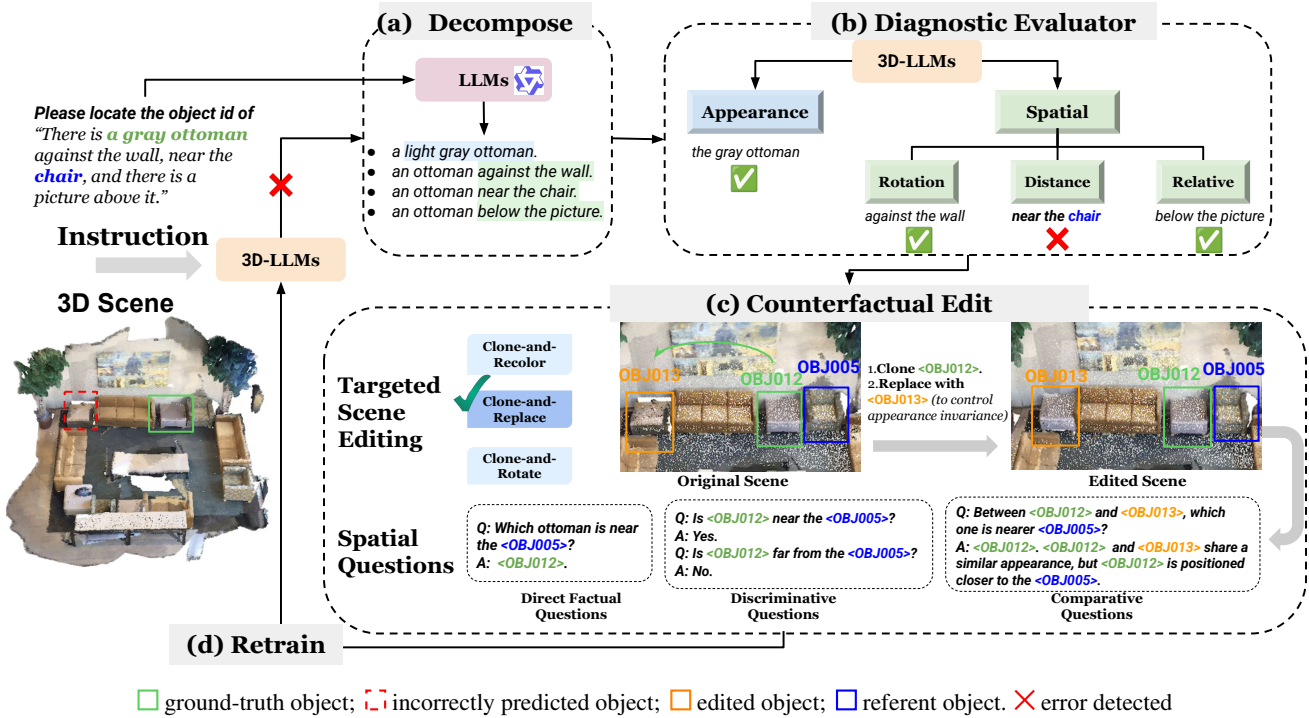

In this section, we introduce our approach, which systematically identifies and corrects grounding errors through targeted counterfactual scene editing.
As illustrated in~\cref{fig:main_fig}, our framework is structured as follows: we first describe our baseline model (\cref{sec:base_model}). Next, we present the core components of our pipeline, beginning with the decomposition of complex instructions into simpler, atomic sub-descriptions, each capturing a distinct semantic factor (\cref{sec:decomposition}). We then introduce a diagnostic evaluator that pinpoints the model’s specific reasoning failures related to these semantic factors (\cref{sec:diagnostic_evaluator}). Then, we detail our scene-editing strategy (\cref{editing}), which generates controlled counterfactual 3D scenes along with aligned question-answer pairs. Finally, we describe how the base model is iteratively fine-tuned using the collected counterfactual data (\cref{sec:retrain}), progressively enhancing its grounding accuracy.

\subsection{Base 3D-LLM Model}
\label{sec:base_model}
We build our pipeline upon a 3D-LLM, denoted as $M_{base}$, which serves as our initial grounding model. Specifically,
we utilize ChatScene~\cite{huang2024chat} as the backbone. In the grounding task, the base model receives inputs consisting of 3D object points $S$ with color and instance segmentation obtained by a 3D segmentor~\cite{Schult2022Mask3DMT}, along with a natural language instruction $T$ describing a target object. 
The model processes these inputs to produce a prediction $O_{pred}$ (a 3D bounding box) that localizes the object specified by the instruction:
\begin{equation}
 O_{pred} = M_{base}(S, T). 
\end{equation}
During training-time supervision construction, the prediction $O_{pred}$ is evaluated against standard dataset annotations to attribute grounding failures at the predicate level. DEER-3D does not assume additional labeling beyond existing training annotations and uses them solely as a deterministic failure attribution signal.  We detail our error-driven analysis framework in the following sections.

\subsection{Instruction Decomposition}
\label{sec:decomposition}
To construct predicate-level supervision signals, we first identify training examples where the base model shows grounding failure under the standard supervised setting.
For these identified cases (i.e., incorrect $O_{pred}$), we decompose complex, free-form instructions $T$ into their atomic components (sub-instructions). As shown in~\cref{fig:main_fig}(a), we employ a large language model (\textit{e.g.}, Qwen3~\cite{yang2025qwen3}) to systematically parse $T$ into a set of atomic predicates, denoted as $\mathbf{P} = \{p_1, p_2, \cdots, p_n\}$. 
This decomposition is crucial, as it isolates potential error sources to individual semantic elements of the original instruction. For example, given the compositional instruction $T$ = \textit{``the brown couch against the wall"}, the parser typically outputs atomic predicates capturing distinct semantic components such as object appearance and spatial relations (e.g., $p_1$ = \textit{``the brown couch"}, $p_2$ = \textit{``the couch is against the wall"}).
The set of atomic predicates $\mathbf{P}$ is then forwarded to our diagnostic evaluator, introduced as follows. Detailed prompts and a manual quality analysis are provided in the Appendix.

\subsection{Diagnostic Evaluator}
\label{sec:diagnostic_evaluator}
Given the set of atomic predicates $\mathbf{P} = \{p_1, ..., p_n\}$, the goal of the diagnostic evaluator is to pinpoint which specific predicate $p_i$ the model failed to ground~(shown in~\cref{fig:main_fig}(b)). From empirical observations, we identified two primary types of visual grounding errors. A detailed analysis of these error categories and their distributions is provided in the Appendix, where we show that these two types collectively account for over +75\% of all errors in our experiments.
We categorize  primary error types as follows:
\begin{itemize}
\item \textbf{Appearance Grounding}: Descriptions related to visual appearance, particularly textures and colors (\textit{e.g., ``brown/wooden couch''}).
\item \textbf{Spatial Grounding}: Descriptions specifying the spatial context, including \textbf{orientation} (\textit{e.g.}, ``\textit{against the wall}"), \textbf{distance} (\textit{e.g.}, ``\textit{near the ottoman}"), or \textbf{relative positioning} (\textit{e.g.}, ``\textit{below the picture}"). 
\end{itemize} 

Importantly, we do not prompt the model to perform self-diagnosis, as current 3D-LLMs lack reliable self-explanation capabilities. Instead, error attribution is conducted through an externally deterministic procedure. The assignment of each sub-instruction $p_i$ to predefined error categories is performed via a rule-based matching mechanism, using a dictionary of appearance and spatial keywords collected from the training corpus (e.g., ``red/yellow'' for appearance, ``against/facing'' for orientation). A parsed sub-instruction is accepted only if it explicitly contains a valid keyword together with the target object's semantic label; otherwise, it is conservatively discarded to prevent noisy edits. Each filtered predicate $p_i$ is then independently queried against the base model $M_{base}$ to obtain a set of candidate object IDs $\{O_{cand}\}$. We determine grounding failure by comparing the predicted candidate set with the annotated target object provided by standard training annotations. 
To assess the reliability of this deterministic attribution protocol, we manually inspect 200 randomly sampled failure cases and find that 85\% of attributed predicates are consistent with human judgment (see Appendix for details).

\begin{figure*}[t]
    \centering
    \includegraphics[width=\linewidth]{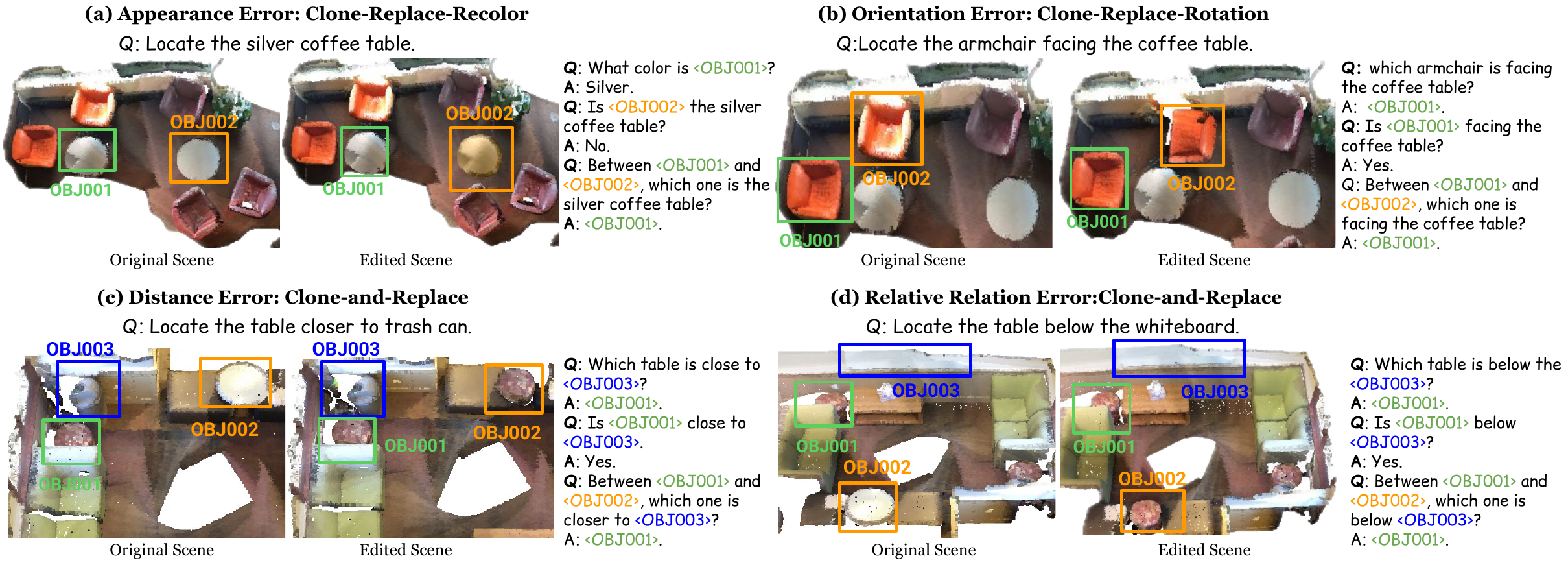}
    \small
    \textcolor{gtgreen}{\tikz{\draw[line width=0.8pt, color=gtgreen] (0,0) rectangle (0.25,0.25);}} targeted;\,
    \tikz{\draw[line width=0.8pt, color=orange] (0,0) rectangle (0.25,0.25);} edited;\,
    \tikz{\draw[line width=0.8pt, color=blue] (0,0) rectangle (0.25,0.25);} referent.
    \caption{
    Examples of targeted visual edits for different grounding errors. Each edit generates aligned question–answer pairs, precisely supervising the model's visual grounding.
    }
    \label{fig:edit example}
\end{figure*}

\subsection{Predicate-Guided Counterfactual Intervention}
\label{editing}
Our method generates controlled 3D counterfactual scenes through predicate-targeted geometric intervention. 
The key principle is to construct minimal intervention pairs that differ along exactly one semantic predicate while preserving all other geometric factors, enabling training supervision primarily constrained to the diagnosed semantic predicate.
Specifically, as shown in~\cref{fig:main_fig}(c), when a visual grounding error is detected, our goal is to isolate the semantic factor responsible for the error while keeping all other visual and geometric properties constant.
To achieve this, we construct \textbf{controlled counterfactual scenes} that differ from the original scene along only one visual predicate. This enables controlled training supervision so that the model must correctly attend to the changed attribute to succeed. 
Moreover, we apply basic heuristic sanity checks (\textit{e.g.}, bounding box collision and floating detection) during editing. While these heuristics do not simulate complex physical interactions, our human evaluation on a sampled subset confirms that the majority (87\%) of the edited scenes remain visually plausible (see more details in Appendix), providing strong, targeted semantic contrast for debiasing.the 
Our editing framework follows a unified \textbf{Clone–Replace–Modify} procedure:

\begin{itemize}
\item \textbf{Clone:} Let $o_{target}$ be the target object. We first create $o_{clone}$, a perfect duplicate of $o_{target}$ with identical geometry, material, and texture properties.
\item \textbf{Replace:} Let $o_d$ be a distractor object, strategically selected based on the diagnosed error type to create a challenging counterfactual example. We remove $o_d$ from the scene and place $o_{clone}$ at the exact 3D position, $T_d$, previously occupied by $o_d$. Formally, this operation is simply:
\[
o_{\texttt{clone}}^{\texttt{relocated}} = \texttt{Translate}(o_{\texttt{clone}}, T_d),
\]
\item \textbf{Modify:} We apply a predicate-specific modification to the newly placed $o_{clone}$. This modification directly corresponds to the diagnosed error type that caused the failure.
\end{itemize}
This unified structure allows the same editing principle to address different visual error categories. In the following, we describe the specific editing operations designed for each visual error type.

\subsubsection{Error-Specific Edits.}
\label{Error-Specific Edits}
While our unified Clone–Replace–Modify framework provides a general mechanism for controlled counterfactual editing, the \textbf{Replace} and the \textbf{Modify} steps vary according to the semantic predicate causing the grounding error.
We provided an example of targeted visual editing for different grounding errors in~\cref{fig:edit example}.

\begin{itemize}
\item \textbf{Clone–Replace–Recolor (CR-Rec) for Appearance Errors.} For appearance errors (e.g., incorrect color identification), we first select a nearby distractor object $o_d$ that shares the same semantic category as the target object $o_{target}$ (\textit{e.g.}, coffee table $<$\texttt{OBJ002}$>$ in~\cref{fig:edit example}(a)).
After cloning and replacing as described above, we recolor the cloned object $o_{clone}$ to a perceptually contrasting hue while preserving its original shading and illumination properties. Formally, the edited object $\tilde{o}$ is obtained by sequentially translating and recoloring:
\[
\tilde{o} = \texttt{Recolor} \!\big(\texttt{Translate}(o_{\texttt{clone}}, T_d),\, c^*\big),
\]
where $T_d$ is the original 3D position of the distractor, and $c^*$ is the selected contrasting hue that maximizes perceptual difference from the targeted color in CIELAB space. Formally,
$c^* = \arg\max_{c \in \mathcal{C}} \|\texttt{Lab}(o_{target}) - \texttt{Lab}(c)\|_2$,
where $\texttt{Lab}(o_{target})$ denotes the original object's mean color in CIELAB space, and $c$ represents a candidate color from a predefined set of distinct candidate hues (more details are provided in Appendix).the 
This controlled recoloring provides two visually identical objects differing in color, enabling precise supervision signals for appearance grounding.

\item\textbf{Clone–Replace–Rotate (CR-Rot) for Orientation Errors.}  
For grounding errors involving orientation predicates (e.g., \textit{``facing the table''}, \textit{``against the wall''}), we first select a nearby distractor object $o_d$ (e.g., the armchair $<$\texttt{OBJ002}$>$ in~\cref{fig:edit example}(b)) from the same semantic category as the target object $o_{target}$. After cloning and replacing as previously described, we apply a controlled rotation to the cloned object $o_{clone}$. Formally, the final edited object $\tilde{o}$ is obtained by sequentially translating and rotating:
\[
\tilde{o} = \texttt{Rotate}~\!\big(\texttt{Translate}(o_{\texttt{clone}}, T_d),\, R_{\mathbf{y}}(\theta)\big),
\]
where $\mathbf{y}$ is the vertical rotation axis and $R_{\mathbf{y}}(\theta)$ denotes a rotation matrix around the Y-axis by angle $\theta$. In our experiments, the rotation angle $\theta$ is selected from a small predefined set $\{\pm45^{\circ}, \pm90^{\circ}\}$ to produce meaningful yet visually consistent orientation changes.
This unified formulation explicitly integrates spatial relocation and orientation modification, yielding two visually identical scenes differing only in object orientation, thereby providing precise supervision signals for directional grounding.

\item\textbf{Clone–and–Replace (CR) for Distance Errors.}  
Distance grounding errors involve a \textbf{referent object} (e.g., the transh can  $<$\texttt{OBJ003}$>$ in~\cref{fig:edit example}(c)), which defines the spatial relationship of interest. 
To identify this referent, we first filter all surrounding objects sharing the same semantic label as the mentioned category (e.g., \textit{``tran can''} instances). 
The specific referent object $o_{ref}$ is then selected according to the spatial predicate in the instruction:
For \textit{``near''} relations, we select the nearest instance to the target object as $o_{ref}$; For \textit{``far''} relations, we select the farthest instance within the same category as $o_{ref}$.
After determining the referent $o_{ref}$, we select a distractor $o_d$ of the same category as the target object $o_{target}$ to construct the counterfactual pair. 
The distractor selection follows the relational constraint:
For a \textit{``near''} predicate, $o_d$ must be farther from $o_{ref}$ than the target object, and vice versa for a \textit{``far''} predicate.
Finally, we replace the selected distractor with the cloned target object.

\item\textbf{Clone–and–Replace (CR) for Relative Relation Errors.}  
We mainly focus on vertical relations (e.g., \textit{“above”}, \textit{“below”}), since horizontal relative-position annotations in ScanNet (e.g., \textit{“left”}, \textit{“right”}) are frequently noisy and lack consistent labeling~\cite{huang2025unveiling}.
Similar to the CR strategy for distance errors, we first identify the referent object that defines the relation in the instruction (e.g., the whiteboard $<$\texttt{OBJ003}$>$ in~\cref{fig:edit example}(d)). 
After identifying $o_{ref}$, we select a distractor object $o_d$ of the same semantic category as the target object $o_{target}$ (e.g., table $<$\texttt{OBJ002}$>$ in~\cref{fig:edit example}(d)) that violates the intended vertical relation.  
For instance, when the instruction specifies ``below'', we select a distractor that is not located below the referent, ensuring a meaningful counterfactual spatial signal.
\end{itemize}

\subsubsection{QA Generation.}

After constructing counterfactual scenes containing both the target object $o_{target}$ and its edited object $\tilde{o}$, we generate new aligned QA pairs to explicitly target the model's original grounding failure.  
Each counterfactual scene is paired with around 5-6 QA examples with increasing reasoning complexity, categorized as follows:

\begin{itemize}
    \item \textbf{Direct Factual (Perception).}  
    Simple open-ended queries assess the model’s perception of each object’s attribute.  
    \textit{Example:}  
    Q: “What color is $o_{target}$?” $\rightarrow$ A: “Light green.”  
    Q: “What color is $\tilde{o}$ ?” $\rightarrow$ A: “Pink.”

    \item \textbf{Discriminative (Verification).}  
    Yes/no questions verify whether the model correctly understands the failed predicate for each object.  
    \textit{Example:}  
    Q: “Is $o_{target}$ light green?” $\rightarrow$ A: “Yes.”  
    Q: “Is $\tilde{o}$  light green?” $\rightarrow$ A: “No.”

    \item \textbf{Comparative (Reasoning).}  
    Each answer identifies the correct object and explicitly states the rationale behind the choice by highlighting differences.
    \textit{Example:}  
    Q: “Between $o_{target}$ and $\tilde{o}$ , which one is light green?” $\rightarrow$ A: “$o_{target}$. $o_{target}$ shows a light tone on its surfaces, while $o_{target}$ appears gold, so their colors differ visibly.” 
\end{itemize}
This mixture of QA types provides comprehensive supervision signals for each counterfactual edit, targeting multiple levels of reasoning difficulty.

\subsection{Iterative Re-training}
\label{sec:retrain}
The re-training phase shown in~\cref{fig:main_fig}(d) completes the \textsc{DEER}-3D loop. Specifically, we combine the original training data with newly generated counterfactual scenes and QA pairs. The base model $M_{base}$ is fine-tuned on this combined dataset to obtain a refined model with improved grounding performance on the previously identified errors. 
The refined model is subsequently re-evaluated under the same deterministic attribution protocol to identify remaining predicate-specific failures. Additional counterfactual training supervision is constructed only for newly identified semantic bottlenecks.

%% file: sec/4_experiment.tex
\section{Experiments}

\subsection{Setup} 

\noindent\textbf{Dataset.} We evaluate our method on standard 3D referring expression grounding benchmark: ScanRefer~\cite{chen2020scanrefer} and Multi3DRefer~\cite{zhang2023multi3drefer}. ScanRefer provides single-object referring expressions, testing a model’s ability to localize fine-grained visual and spatial attributes.
Multi3DRefer builds upon this by introducing multi-object references and relational reasoning.

\noindent\textbf{Baseline.} We adopt two backbone variants of the state-of-the-art \textbf{Chat-Scene} baseline~\cite{huang2024chat}: (1) \textbf{Chat-Scene (3D-only)}, which utilizes textual instructions and 3D point clouds without 2D visual inputs, and (2) \textbf{Chat-Scene}, the original version that additionally incorporates complementary 2D multi-view images. This setup enables us to assess our approach under different modality conditions.

\noindent\textbf{Evaluation Metrics.} We adopt standard evaluation metrics specific to each benchmark. For grounding tasks on ScanRefer~\cite{chen2020scanrefer} and Multi3DRefer~\cite{zhang2023multi3drefer}, we report grounding accuracy (Acc) at Intersection over Union (IoU) thresholds of 0.25 (Acc@0.25) and 0.5 (Acc@0.5), where predictions are considered correct if their IoU with anchor annotations exceeds the corresponding threshold. For multi-object grounding on Multi3DRefer~\cite{zhang2023multi3drefer}, we report F1 scores at these IoU thresholds (F1@0.25, F1@0.5).

\input{tables/main_grounding}

\noindent\textbf{Implementation Details.}
In Round 1, we fine-tune the baseline model using the original training data plus ~10k counterfactual scenes and ~60k QA pairs.
In Round 2, we use the Round 1 model to identify new failure cases, generate an additional ~3k counterfactual scenes and ~18k QA pairs, and further fine-tune the model to obtain the Round 2 model.
For all fine-tuning stages, we train for $5$ epochs using the AdamW optimizer. The learning rate is $3e^{-5}$, with a weight decay of $0.01$ and a global batch size of $32$. All experiments are conducted on 4 NVIDIA H100 (80 GB) GPUs.
More details are provided in Appendix.

\subsection{Experimental Results}

\noindent\textbf{Grounding Performance.}
As shown in~\cref{tab:grounding_comparison}, our approach consistently outperforms prior models across both 3D grounding benchmarks. While traditional expert models rely on task-specific architectures, our goal is to enhance \textit{LLM-based} models that jointly reason over language and 3D scenes.
Results demonstrate significant performance gains, particularly notable in the purely 3D setting, where our counterfactual augmentation delivers improvements of up to around $+5\%$ in the first round of augmentation alone. This highlights our method's effectiveness in directly enhancing spatial and attribute grounding from 3D signals.
Moreover, when combined with the richer multimodal inputs provided by Chat-Scene, our approach achieves further improvements, establishing a new state-of-the-art performance across both datasets. 
We also provide per–subset results for these benchmarks in Table~\ref{tab:more detailed results on Scanrefer} and Table~\ref{tab:multirefer-pertask}, respectively.
In particular, we observe that the improvements of DEER-3D are particularly obvious in the Multiple subset of ScanRefer and multi-task in MultiRefer, where multiple objects of the same category appear in the scene. 
This setting is more challenging because models must rely on fine-grained spatial relations or attribute cues to disambiguate between similar objects, rather than exploiting dataset-level biases. The results further support our motivation that targeted counterfactual supervision is particularly beneficial in scenarios where grounding requires resolving object ambiguity.

\begin{table}[t]
\centering
\caption{Per-subset performance comparison on ScanRefer. Unique and Multiple denote queries referring to object categories with a single and multiple instances.}
\label{tab:more detailed results on Scanrefer}
\resizebox{0.85\linewidth}{!}{
\begin{tabular}{lcccccc}
\toprule
\multirow{2}{*}{Method} &
\multicolumn{2}{c}{Unique} & 
\multicolumn{2}{c}{Multiple} & 
\multicolumn{2}{c}{Overall} \\
\cmidrule(lr){2-3} \cmidrule(lr){4-5} \cmidrule(lr){6-7}
 & Acc@0.25 & Acc@0.5 & Acc@0.25 & Acc@0.5 & Acc@0.25 & Acc@0.5 \\
\midrule
Chat-scene~\cite{huang2024chat} & $\mathbf{89.6}$ & $\mathbf{82.5}$ & $47.8$ & $42.9$ & $55.5$ & $50.2$ \\
Ours & $88.6$ & $81.9$ & $\mathbf{50.8}$ & $\mathbf{45.8}$ & $\mathbf{57.8}$ & $\mathbf{52.3}$ \\
\bottomrule
\end{tabular}}
\end{table}

\begin{table*}[t]
\centering
\caption{Per-subset performance comparison on MultiRefer. ZT: Zero-shot, ST: Single-task, MT: Multi-task, D: Distractor. Best results are bolded.}
\label{tab:multirefer-pertask}
\small
\setlength{\tabcolsep}{4pt}
\resizebox{\textwidth}{!}{
\begin{tabular}{lcc|cc|cc|cc|cc}
\toprule
\multirow{2}{*}{Method}
& \multicolumn{1}{c|}{ZT w/o D}
& \multicolumn{1}{c|}{ZT w/ D}
& \multicolumn{2}{c|}{ST w/o D}
& \multicolumn{2}{c|}{ST w/ D}
& \multicolumn{2}{c|}{MT}
& \multicolumn{2}{c}{Overall} \\

\cmidrule(lr){2-3}
\cmidrule(lr){4-5}
\cmidrule(lr){6-7}
\cmidrule(lr){8-9}
\cmidrule(lr){10-11}

& F1 & F1
& F1@0.25 & F1@0.5
& F1@0.25 & F1@0.5
& F1@0.25 & F1@0.5
& F1@0.25 & F1@0.5 \\

\midrule
3DVG-Trans+~\cite{Zhao20213DVGTransformerRM}
&$87.1$&$45.8$&--&$27.5$&--&$16.7$&--&$26.5$&--&$25.5$\\

D3Net~\cite{Chen2021D3NetAS}
&$81.6$&$32.5$&--&$36.6$&--&$23.3$&--&$35.0$&--&$32.2$\\

3DJCG~\cite{Cai20223DJCGAU}
&$\mathbf{94.1}$&$66.9$&--&$26.0$&--&$16.7$&--&$26.2$&--&$26.6$\\

M3DRef-CLIP~\cite{zhang2023multi3drefer}
&$81.8$&$39.4$&$53.5$&$47.8$&$34.6$&$30.6$&$43.6$&$37.9$&$42.8$&$38.4$\\

\midrule

Chat-Scene~\cite{huang2024chat}
&$90.3$&$62.6$&$82.9$&$\mathbf{75.9}$&$49.1$&$44.5$&$45.7$&$41.1$&$57.1$&$52.4$\\

Ours
&$92.6$&$\mathbf{69.3}$&$82.1$&$75.6$&$\mathbf{52.6}$&$\mathbf{48.4}$
&$\mathbf{50.0}$&$\mathbf{46.0}$
&$\mathbf{60.0}$&$\mathbf{55.8}$\\

\bottomrule
\end{tabular}}
\end{table*}


\begin{wraptable}{r}{0.4\linewidth}
\centering
\small
\setlength{\tabcolsep}{4pt}
\caption{Cross-backbone generalization on ScanRefer.}
\begin{tabular}{lcc}
\toprule
Method & Acc@0.25 & Acc@0.5 \\
\midrule
LL3DA~\cite{chen2024ll3da} & $25.7$ & $20.8$ \\
Ours  & $\mathbf{29.2}$ & $\mathbf{25.2}$ \\
\bottomrule
\end{tabular}
\label{tab:ll3da_transfer}
\end{wraptable}
\noindent\textbf{Error-Guided Iterative Refinement.} 
We further validate the effectiveness of our error-guided iterative editing loop, as shown in~\cref{tab:grounding_comparison}. Starting from the initial improvement achieved by the first round of counterfactual augmentation (Round 1), we apply a subsequent iteration (Round 2), producing additional performance gains across both benchmarks and baselines. Crucially, these iterative gains are consistent for both purely 3D and multimodal baselines, highlighting that our approach is universally beneficial regardless of input modalities. This clearly demonstrates that our iterative loop is more than just a one-time data enhancement; instead, it serves as a sustainable self-correction mechanism, progressively refining the model by systematically addressing its weaknesses. More detailed analyses of each iterative step are provided in our ablation study.

\noindent\textbf{Cross-Backbone Generalization.}  To evaluate whether DEER-3D generalizes beyond the object-centric Chat-Scene backbone, we additionally apply the counterfactual refinement to 
LL3DA~\cite{chen2024ll3da}, a representative scene-level 3D-LLM that directly operates on entire point clouds and predicts continuous 3D bounding boxes. 
As shown in~\cref{tab:ll3da_transfer}, we observe consistent performance gains in ScanRefer, suggesting that our error-driven counterfactual training supervision transfers across different 3D-LLM input paradigms.

\begin{figure*}[t]
  \centering
  \begin{subfigure}{0.32\linewidth}
    \includegraphics[width=\linewidth]{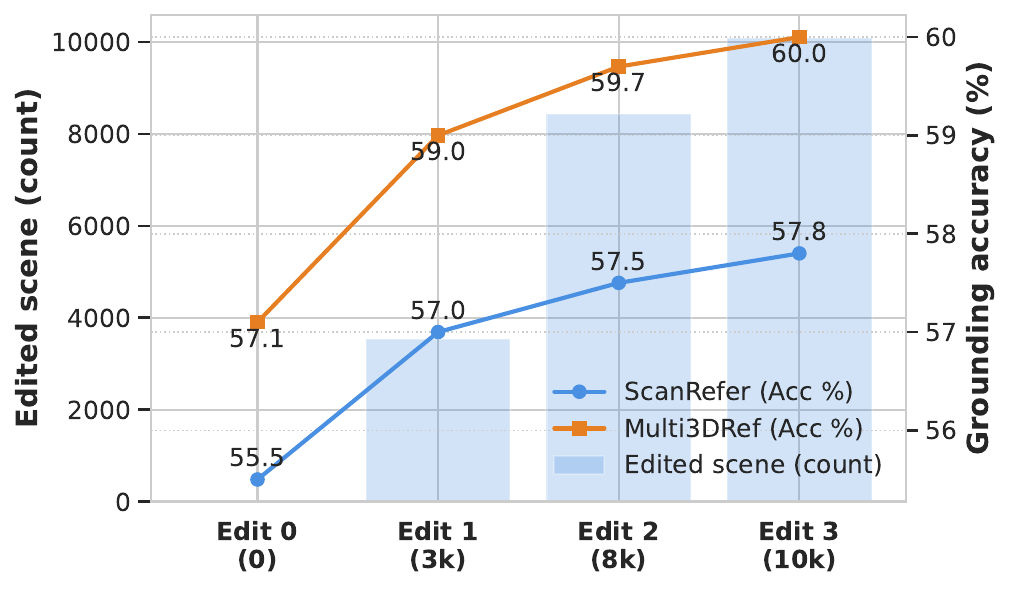}
    \caption{Edit scaling effects.}
    \label{fig:edit_volume}
  \end{subfigure}
  \hfill
  \begin{subfigure}{0.32\linewidth}
    \includegraphics[width=\linewidth]{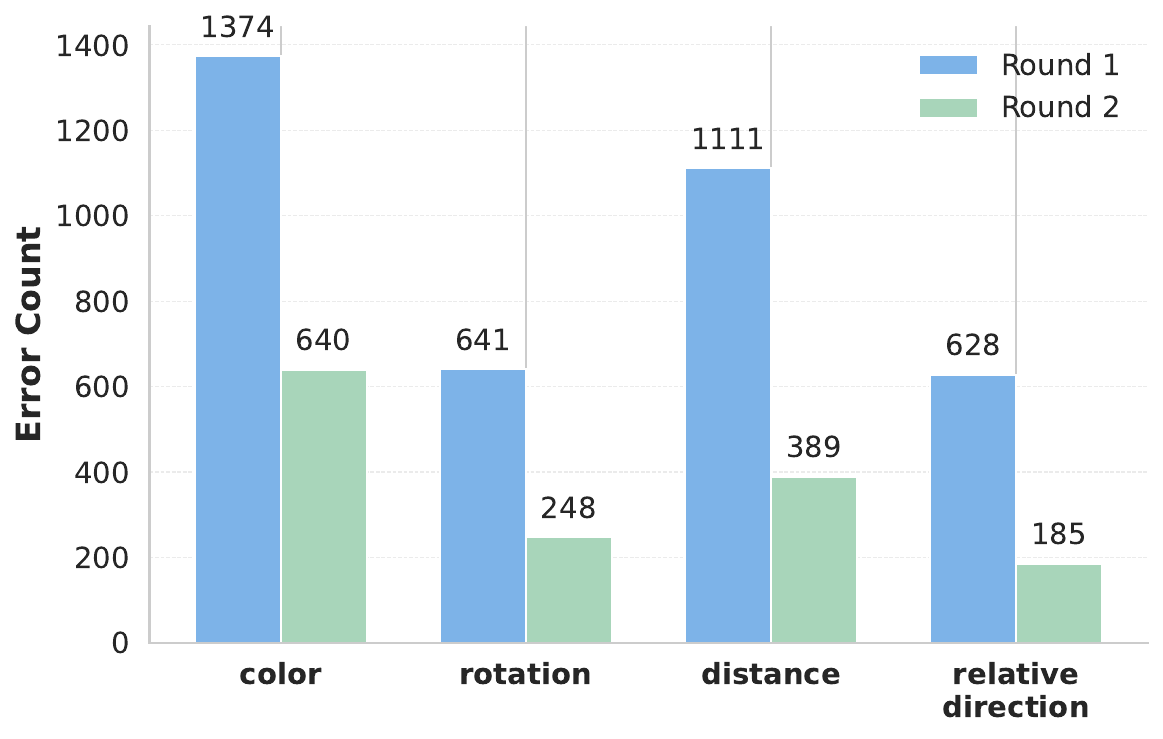}
    \caption{Error distributions.}
    \label{fig:error_counts}
  \end{subfigure}
   \hfill
  \begin{subfigure}{0.32\linewidth}
    \includegraphics[width=\linewidth]{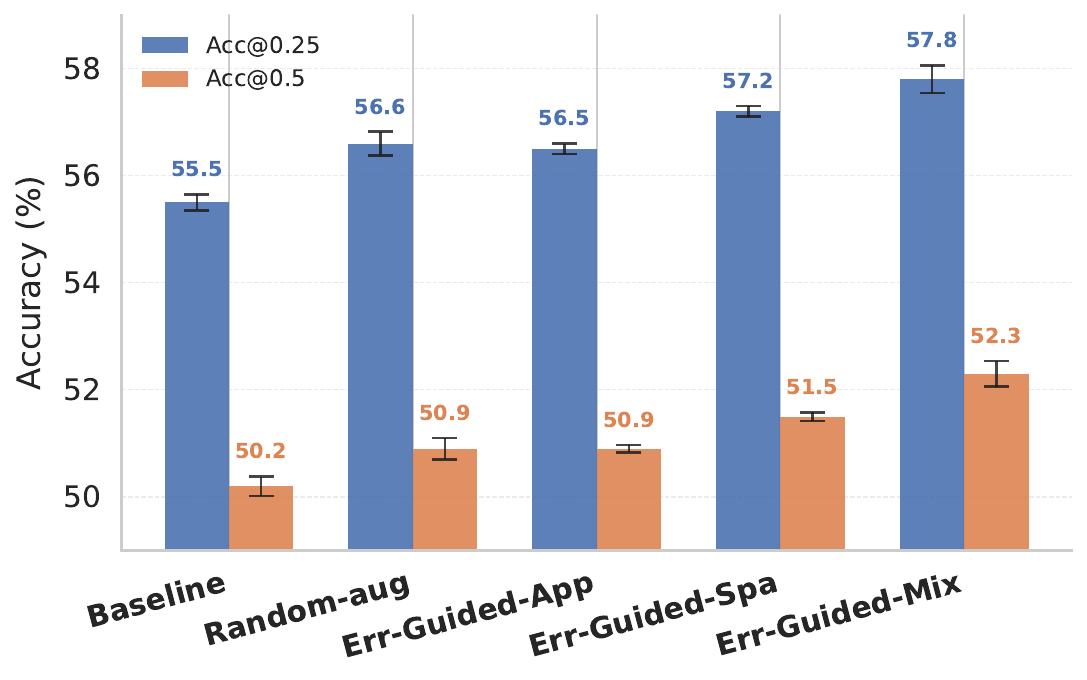}
    \caption{Different edit strategies}
    \label{fig:ablation_edits}
  \end{subfigure}
  \caption{Ablation studies validating our method. (a) Grounding performance scales with the quantity of our counterfactual data. (b) Our iterative loop reduces all targeted error types. (c) Our full Err-Guided-Mix strategy is superior to other strategies.
  } 
  \label{fig:scaling_results}
\end{figure*}

\subsection{Ablation Study}
In this section, we present ablation studies conducted with Chat-Scene with 2D images as our baseline model. More ablations are provided in the Appendix.

\noindent\textbf{Effect of Augmentation Scale.} We analyze how grounding performance scales with the amount of counterfactual data generated by varying the number of edited distractors around each ground-truth object. Specifically, in~\cref{fig:edit_volume}, each increment in the ``Edit" setting (Edit-0 → Edit-3) corresponds to progressively increasing the sampling of distractors per scene (e.g., from 1 to 3, then 5), resulting in a cumulative increase in total edited scenes. Increasing the number of edited distractors per scene (Edit-0→Edit-3) consistently improves performance on both benchmarks. The largest gains occur in the initial scaling stage, while further increments yield diminishing returns, indicating a balance between counterfactual diversity and augmentation efficiency.

\noindent\textbf{Error Distribution Shift Across Refinement Rounds.}  
While~\cref{tab:grounding_comparison} shows overall gains, in~\cref{fig:error_counts}, we further explicitly compare grounding error counts between Round 1 and Round 2 across predicate categories. We observe substantial reductions in errors for all predicate types, clearly indicating that our framework effectively identifies and addresses the model's primary grounding failures, thereby progressively refining its visual and spatial grounding capabilities.

\input{tables/main_understanding}

\noindent\textbf{Ablation on Edit Strategies.}  
To analyze the effectiveness of each edit type within our counterfactual augmentation framework, we conduct an ablation study as shown in~\cref{fig:ablation_edits}.  
Starting from the baseline, a simple Random-aug strategy, which applies edits randomly while keeping the total number of edits equal to that of our full strategy, already yields a modest gain. 
This indicates that generic augmentation alone can slightly enhance model performance.
In contrast, our targeted strategies are far more efficient and effective. Notably, our Err-Guided-App (appearance-only) and Err-Guided-Spa (spatial-only)  achieve a comparable performance despite using significantly less data than the full Random-aug set.
Finally, combining these two targeted edit types into Err-Guided-Mix leads to the highest performance, showing the effectiveness of our counterfactual edits.

\subsection{Other Analysis}

\noindent\textbf{Performance on Captioning and QA.} We further assess the generalization of our grounding-focused augmentation on broader scene understanding tasks, including captioning and question answering (Scan2Cap~\cite{chen2021scan2cap}, ScanQA~\cite{azuma2022scanqa}, and SQA3D~\cite{ma2022sqa3d}). As shown in~\cref{tab:qa_comparison}, our method achieves consistent improvements over the baseline across all benchmarks, indicating that the benefits of our approach extend beyond grounding-specific settings. Detailed metric definitions and additional analysis are provided in the Appendix.

\noindent\textbf{Evaluation of Error-guided Diagnosis without Visual Editing.}
To evaluate the effectiveness of our error-guided diagnostic framework without visual edits, we conduct an experiment using a text-only strategy (``\textsc{DEER}-3D + Text-Aug"). Specifically, we leverage our diagnostic pipeline to pinpoint ambiguous errors, but instead of visually editing the 3D scenes, we generate clearer textual instructions that explicitly disambiguate each failure case. For instance, if the model confuses two visually similar radiators, our framework identifies this ambiguity and generates a more precise instruction (e.g., ``\textit{locate the radiator to the north}"), while leaving the scene unchanged. Compared to random text augmentation, our error-guided text refinement achieves a modest but consistent improvement (55.8\% → 56.4\% Acc@0.25 on ScanRefer), indicating that structured failure-aware training supervision is more effective than unguided augmentation. 
However, the gains remain limited compared to our visual counterfactual editing framework, suggesting that textual clarification alone cannot resolve visually entangled grounding ambiguities and that predicate-level visual intervention plays a crucial role.  Please refer to the Appendix for more details.

\begin{table}[t]
    \centering
    \caption{Experimental results on Beacon3D~\cite{huang2025unveiling}.  “App.”, “Geo.”, and “Spa.” denote appearance, geometry, and spatial relations, respectively. 
    “Overall (Case)” reports average accuracy per case, while “Overall (Obj)” reports average accuracy per object.}
    \label{tab:beacon_results}
    \renewcommand{\arraystretch}{1.15}
    \resizebox{0.7\linewidth}{!}{
    \begin{tabular}{lcccccc}
        \toprule
        \textbf{Model} & \textbf{Class} & \textbf{App.} & \textbf{Geo.} & \textbf{Spa.} & \textbf{Overall (Case)} & \textbf{Overall (Obj)} \\
        \midrule
        Chat-Scene~\cite{huang20253d} & $59.8$ & $59.2$ & $50.2$ & $53.8$ & $59.8$ & $41.0$ \\
        \textbf{Ours} & $\mathbf{61.7}$ & $\mathbf{64.4}$ & $\mathbf{53.4}$ & $\mathbf{55.5}$ & $\mathbf{61.7}$ & $\mathbf{45.7}$\\
        \bottomrule
    \end{tabular}
    }
\end{table}

\noindent\textbf{Evaluation on Human-annotated 3D Grounding.}  
To further validate the real-world reliability of our method, we evaluate on the BEACON-3D benchmark~\cite{huang2025unveiling}, which provides manually verified object and relation annotations for 3D scenes. 
This dataset reflects human judgments of spatial and visual grounding, offering a more rigorous test of perceptual understanding. 
As shown in~\cref{tab:beacon_results}, our model outperforms Chat-Scene across all categories, indicating that our error-driven augmentation not only enhances quantitative performance but also yields more human-aligned grounding behavior.
While DEER-3D demonstrates consistent improvements, we provide a broader discussion on remaining challenges and future directions in the Appendix.

%% file: tables/main_grounding.tex
\begin{table}[t]
\centering
\caption{Performance comparison on 3D grounding benchmarks. Our method consistently outperforms baselines, with iterative re-training further improving results.}
\label{tab:grounding_comparison}
\resizebox{0.75\linewidth}{!}{
\begin{tabular}{lcccc}
\toprule
\multirow{2}{*}{\textbf{Method}} &
\multicolumn{2}{c}{\textbf{ScanRefer}} &
\multicolumn{2}{c}{\textbf{Multi3DRefer}} \\
\cmidrule(lr){2-3} \cmidrule(lr){4-5}
 & Acc@0.25 & Acc@0.5 & F1@0.25 & F1@0.5 \\
\midrule
\multicolumn{5}{l}{\textbf{Expert Models}} \\
\midrule
ScanRefer~\cite{chen2020scanrefer} & $37.3$ & $24.3$ & - & - \\
3DJCG~\cite{cai20223djcg} & $49.6$ & $37.3$ & - & - \\
M3DRef-CLIP~\cite{zhang2023multi3drefer} & $51.9$ & $44.7$ & $42.8$ & $38.4$ \\
3D-VisTA~\cite{zhu20233d} & $50.6$ & $45.5$ & - & - \\
\midrule
\multicolumn{5}{l}{\textbf{LLM-based Models}} \\
\midrule
3D-LLM~\cite{hong20233d} & $30.3$ & - & - & -  \\ 
Chat-3D~\cite{wang2023chat} & $35.9$ & $30.4$ & - & - \\
Chat-3D v2~\cite{huang2024chat} & $42.5$ & $38.4$ & $45.1$ & $41.6$ \\
Grounded 3D-LLM~\cite{chen2024grounded}  & $47.9$ & $ 44.1$ & $45.2$ & $40.6$ \\ 
PQ3D~\cite{zhu2024unifying} & $57.0$ & $51.2$ & - &  $50.1$ \\
3D-LLaVA~\cite{deng20253d} & $51.2$ &  $40.6$ & - & -\\
Inst3D-LLM~\cite{yu2025inst3d} & $57.8$ & $51.6$ & $58.3$ & $45.5$ \\
\midrule
\rowcolor{gray!10} Chat-Scene (w/o 2D)~\cite{huang2024chat} & $41.2$ & $37.4$ & $43.8$ & $40.2$ \\
\textbf{Ours (Round 1)} & 
$45.5$~{\scriptsize\textcolor{blue}{($+4.3$)}} & 
$41.8$~{\scriptsize\textcolor{blue}{($+4.4$)}} &
$48.4$~{\scriptsize\textcolor{blue}{($+4.6$)}} &
$45.1$~{\scriptsize\textcolor{blue}{($+4.9$)}} \\
\textbf{Ours (Round 2)} & 
$47.1$~{\scriptsize\textcolor{blue}{($+5.9$)}} &
$43.1$~{\scriptsize\textcolor{blue}{($+5.7$)}} &
$50.3$~{\scriptsize\textcolor{blue}{($+6.5$)}} &
$47.1$~{\scriptsize\textcolor{blue}{($+6.9$)}} \\
\midrule
\rowcolor{gray!10}Chat-Scene~\cite{huang2024chat} 
& $55.5$ & $50.2$ & $57.1$ & $52.4$ \\
\textbf{Ours (Round 1)} 
& 57.8~{\scriptsize\textcolor{blue}{($+2.3$)}} 
& 52.3~{\scriptsize\textcolor{blue}{($+2.1$)}} 
& 60.0~{\scriptsize\textcolor{blue}{($+2.9$)}} 
& 55.8~{\scriptsize\textcolor{blue}{($+3.4$)}} \\
\textbf{Ours (Round 2)} 
& $\mathbf{58.6}$~{\scriptsize\textcolor{blue}{($+3.1$)}} 
& $\mathbf{53.3}$~{\scriptsize\textcolor{blue}{($+3.1$)}} 
& $\mathbf{61.4}$~{\scriptsize\textcolor{blue}{($+4.3$)}} 
& $\mathbf{56.8}$~{\scriptsize\textcolor{blue}{($+4.4$)}} \\
\bottomrule
\end{tabular}}
\end{table}

%% file: tables/main_understanding.tex
\begin{table}[t]
\centering
\caption{Experimental results on scene understanding tasks. Our method outperforms baselines and shows strong overall capability across general scene understanding tasks.}
\label{tab:qa_comparison}
\resizebox{0.7\linewidth}{!}{
\begin{tabular}{lcccccc}
\toprule
\multirow{2}{*}{\textbf{Method}} &
\multicolumn{2}{c}{\textbf{Scan2Cap}} &
\multicolumn{2}{c}{\textbf{ScanQA}} &
\multicolumn{2}{c}{\textbf{SQA3D}} \\
\cmidrule(lr){2-3} \cmidrule(lr){4-5} \cmidrule(lr){6-7}
 & C@0.5 & B-4@0.5 & C & B-4 & EM & EM-R \\
\midrule
3D-LLM~\cite{hong20233d}  & - & - & $69.4$ & $12.0$ & - & - \\
Chat-3D v2~\cite{wang2023chat} & $63.9$ & $31.8$ & $87.6$ & $14.0$ & $54.7$ & - \\
LL3DA~\cite{chen2024ll3da} & $65.2$ & $36.8$ & $76.8$ & $13.5$ & - & $53.7$ \\
3D-LLaVA~\cite{deng20253d} & $78.7$ & $36.9$ & $92.6$ & - & $54.5$ & - \\
LEO~\cite{huang2023embodied} & $68.4$ & $\mathbf{36.9}$ & $80.0$ & $11.5$ & - & $53.7$ \\
SceneLLM~\cite{fu2024scene} & - & - & $80.0$ & $12.0$ & - & $54.2$ \\
\midrule %
\rowcolor{gray!15} %
Chat-Scene (w/o 2D)~\cite{huang2024chat}  & $64.9$ & $-$ & $80.3$ & $-$ & $53.4$ & - \\
\textbf{Ours} & $72.4$ & $35.0$ & $88.7$ & $14.3$ & $53.6$ & $56.3$\\
\midrule
\rowcolor{gray!15} %
Chat-Scene~\cite{huang2024chat} & $77.1$ & $36.3$ & $87.7$ & $\mathbf{14.3}$ & $54.6$ & $57.5$ \\
\textbf{Ours} & $\mathbf{81.4}$ & $35.8$ & $\mathbf{89.5}$ & $13.1$ & $\mathbf{56.3}$ & $\mathbf{58.9}$ \\
\bottomrule
\end{tabular}}
\setlength{\parskip}{0pt}
\end{table}

%% file: sec/5_conclusion.tex
\section{Conclusion}
We introduce DEER-3D, an error-driven framework that enhances 3D grounding via targeted visual counterfactual editing. \textsc{DEER}-3D diagnoses grounding errors, decomposes them into semantic factors, and generates controlled 3D edits for precise predicate-level supervision. Through iterative refinement, our approach consistently improves grounding performance. Comprehensive experiments and analysis further demonstrate that error-guided scene editing effectively strengthens grounding capabilities in 3D-LLMs.

\section*{Acknowledgment}
We thank Jaemin Cho, Elias Stengel-Eskin, and Zaid
Khan for their helpful feedback. This work was supported by NSF-AI Engage Institute DRL-2112635, ARO Award W911NF2110220, ONR Grant N00014-23-1-2356, DARPA ECOLE Program No. HR00112390060, and a Capital One Research Award. The views contained in this article are those of the authors and not of the funding
agency.

%% file: sec/X_suppl.tex
\clearpage
\appendix
\setcounter{page}{1}
\appendix

\section*{Appendix}

\section{More Details about Deer3D Framework}
\label{more method discussion}
In this section, we provide additional implementation details about the DEER-3D framework. We elaborate on the specific prompts used for our Decomposition module (Sec.~\ref{sec:decomposition}), the full statistical analysis of the instruction types that justifies our focus on Appearance and Spatial categories (Sec.~\ref{sec:diagnostic_evaluator}), and the complete color palette $\mathcal{C}$ used for our Recolor operation (Sec.~\ref{Error-Specific Edits}).

\subsection{Backbones}
\label{sec:supp_backbone}
In our experiments, we employ Chat-Scene as the primary 3D-LLM backbone, which integrates multi-modal inputs including 3D point clouds, flexible 2D multi-view images, and textual instructions. 
Specifically, for the 3D modality, we utilize object-centric features extracted by the pre-trained Uni3D encoder~\cite{Zhou2023Uni3DEU}, which captures both geometric structure and semantic information from the point cloud. 
For the 2D modality, we employ the pre-trained DINOv2 encoder~\cite{Oquab2023DINOv2LR} to extract dense visual features from multi-view images. 
To align these visual signals with the language space, trainable Multi-Layer Perceptrons (MLPs) project both the Uni3D and DINOv2 features into the token embedding space of the Large Language Model. The core language backbone is Vicuna-7B-v1.5~\cite{chiang2023vicuna}, which takes the concatenated sequence of projected visual tokens and text instruction tokens to generate the final response.

To evaluate the generality of our approach across different 3D-LLM architectures, we additionally conduct experiments on LL3DA~\cite{chen2024ll3da}. 
Unlike Chat-Scene, which relies on object-centric representations, LL3DA directly operates on the entire scene-level point cloud and predicts continuous 3D bounding boxes through a detection-style architecture. 
This design allows LL3DA to reason over global scene geometry without relying on pre-segmented object features.

\subsection{Prompt for Decompose Instruction}
\label{Prompt for Decompose Instruction}
As detailed in Sec.~\ref{sec:decomposition} of the main paper, our DEER-3D framework begins by decomposing the full instruction into a set of general, atomic predicates. We use a large language model (e.g., Qwen3) for this task.To ensure the generated sub-instructions are consistent, unambiguous, and suitable for independent evaluation, we use a structured prompt. This prompt instructs the LLM to separate the instruction into a list of sentences, each focusing on a single semantic perspective (e.g., attribute, spatial relation). The prompt enforces several key constraints to maintain quality: it explicitly forbids the use of ambiguous referring expressions (like ``it" or ``them") and requires that all output sentences maintain the same, explicit subject (e.g., ``The table is brown," ``The table is near the wall").

\subsection{Human Quality Analysis}
\label{human quality analysis}

To evaluate the reliability of our pipeline, we conduct human quality analysis from two perspectives: 
(1) the correctness of the LLM-based instruction decomposition and 
(2) the visual plausibility of the edited 3D scenes.

\noindent\textbf{Instruction Decomposition Quality.} 
To assess the quality of the decomposition, we randomly sample 100 instructions and perform a human evaluation.
We do not observe hallucinated objects or attributes in the generated sub-instructions. 
The only observed errors include: 
(1) \textit{under-splitting} of multi-attribute instructions (3 cases), and 
(2) \textit{reference re-association} across sub-instructions (6 cases). 
Importantly, these cases do not lead to incorrect scene edits. 
If a sub-instruction cannot be unambiguously matched to a predefined error type or a valid target object, it is filtered out by our rule-based gate, resulting in a conservative \emph{no-edit} decision rather than an erroneous edit.

\noindent\textbf{Predicate Classification.} To categorize the predicate type of each sub-instruction, we apply a template-based classification procedure using predefined token dictionaries. 
Specifically, we construct token sets for different predicate categories, including visual attributes and spatial relations (see~\cref{predicate category}). 
Each sub-instruction is assigned to a predicate category based on token matching. 
This deterministic procedure enables us to identify which attribute or spatial relation is responsible for a grounding failure and to apply the corresponding scene-editing operation.
To assess the reliability of this rule-based classification protocol, we manually inspect 200 randomly sampled failure cases. 
Human annotators compare the automatically assigned predicate category with the correct predicate implied by the instruction. 
We find that 85\% of the automatically classified predicates are consistent with human judgment, suggesting that the template-based classification provides a reliable signal for downstream editing decisions.

\begin{table}[t]
\centering
\caption{Vocabulary for predicate classification in the diagnostic evaluator. }
\label{predicate category}
\small
\begin{tabular}{ll}
\toprule
\textbf{Predicate Category} & \textbf{Example Tokens} \\
\midrule
Color / Appearance & white, black, red, blue, brown, etc. \\
Relative Direction & above, below, under, on top, etc.  \\
Distance Relation & next to, near, far from, closest to, etc.  \\
Orientation / Rotation & facing, opposite to, against, etc.  \\
\bottomrule
\end{tabular}
\end{table}

\noindent\textbf{Edited Scene Plausibility.}
To ensure realistic scene modifications, we enforce several sanity constraints during editing, including collision detection, floating-object checks, selection of similar-size distractors, and color perturbations in a perceptually aligned color space (CIELAB). 
We further conduct a human evaluation on 60 randomly sampled edited scenes. 
Annotators are asked to judge whether the edited scene appears visually plausible without introducing obvious geometric or physical inconsistencies. 
Results show that the majority of edits (87\%) are considered visually plausible by human evaluators, suggesting that our editing procedure preserves realistic scene configurations in most cases.
These results indicate that the proposed editing pipeline introduces limited visual artifacts while maintaining the semantic correctness of the edited strategies.

\begin{wrapfigure}{r}{0.42\linewidth}
\centering
\includegraphics[width=\linewidth]{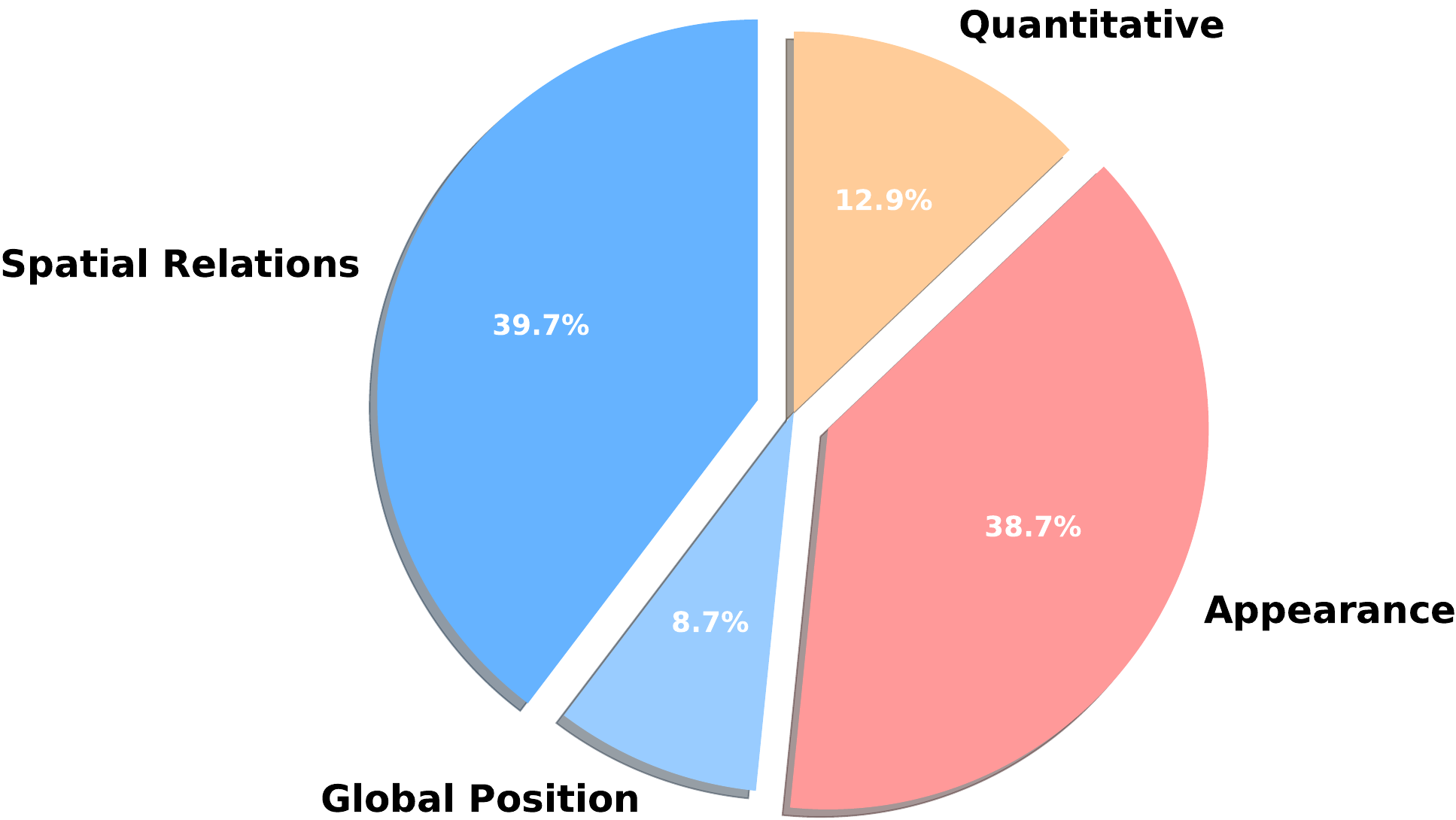}
\caption{Distribution of semantic predicate types.}
\label{fig:error_analysis_pie_chart}
\end{wrapfigure}

\subsection{Analysis of Error Distribution}
\label{Error Distribution Analysis}
In our methodology (Sec.~\ref{sec:diagnostic_evaluator}), we state that our framework focuses on Appearance and Spatial errors. This decision is data-driven, based on a statistical analysis of the instruction types present in the training data.
To quantify the most common semantic components, we randomly sample around 300 instructions from the training dataset. We then ran our Decomposition module (Sec.~\ref{sec:decomposition}) on this sample to parse each instruction into its constituent atomic predicates (around 1500 sub-instructions).
The distribution of predicate types within this representative sample is presented in~\cref{fig:error_analysis_pie_chart}. The analysis reveals that the instructions are not uniformly distributed; they are overwhelmingly concentrated in two primary categories: Spatial Relations (39.6\%) and Appearance (38.6\%).
Together, these two categories conclusively account for 78.2\% of all semantic predicates found in our analysis. Assuming this large sample is representative of the full dataset, this provides clear empirical evidence that Spatial and Appearance are the most dominant semantic components in the data. Therefore, our DEER-3D framework is justifiably prioritized to focus its diagnostic and editing efforts on these two categories.

\begin{tcolorbox}[
    colback=gray!5,
    colframe=gray!50,
    boxrule=0.7pt,
    arc=2pt,
    left=5pt,right=5pt,top=5pt,bottom=5pt,
    title={\textbf{Prompt for Decompose Instruction}}
]
\small
\texttt{Given an instruction sentence, separate it into a list of sentences where each focuses on only one perspective, such as spatial relationship, attribute, or quantity.}
\texttt{Important constraints:}
\begin{itemize}
    \item \texttt{Do NOT use referring expressions like "it" or "them". Explicitly state what these refer to.}
    \item \texttt{Ensure that all output sentences use the SAME subject throughout.}
    \item \texttt{If other objects are involved (e.g., "a computer is placed on top of it"), rephrase so the original subject remains the focus (e.g., "The table has a computer placed on top of it").}
\end{itemize}
\texttt{Example:}\\
\texttt{Input: "There is a brown wooden table, a computer is placed on top of it."}\\
\texttt{Output: ["There is a brown table", "There is a wooden table", "The table has a computer placed on top of it."]}
\medskip
\end{tcolorbox}

\subsection{Edit Strategy for Recolor}
\label{edit strategy for Recolor}

In Sec.~\ref{Error-Specific Edits}, our Recolor operation identifies the most perceptually distinct color $c^*$ by maximizing its distance from a candidate set $\mathcal{C}$. To ensure that this process is both reproducible and perceptually meaningful, we define $\mathcal{C}$ as a predefined palette of 19 commonly used color prototypes.  
We operate in the CIELAB color space due to its perceptual uniformity, which makes Euclidean distances better aligned with human color perception. Our full LAB dictionary, and the candidate set $\mathcal{C}$ is listed below, mapping each color name to its corresponding (L*, a*, b*) values.

\begin{table}[h]
\centering
\small
\caption{The predefined set of color ($\mathcal{C}$) used for recolor operation.}
\resizebox{0.85\linewidth}{!}{
\begin{tabular}{ll|ll}
\toprule
\textbf{Color Name} & \textbf{CIELAB (L, a, b)} & \textbf{Color Name} & \textbf{CIELAB (L, a, b)} \\
\midrule
white       & (90, 0, 0)     & green      & (55, -35, 35) \\
black       & (12, 0, 0)     & turquoise  & (70, -35, -10) \\
gray        & (55, 0, 0)     & blue       & (40, 5, -55) \\
beige       & (68, 5, 18)    & dark blue  & (30, 5, -45) \\
tan         & (62, 10, 22)   & purple     & (45, 45, -25) \\
brown       & (35, 18, 28)   & pink       & (70, 30, 10) \\
dark brown  & (28, 14, 20)   & violet     & (50, 35, -35) \\
red         & (45, 60, 30)   & silver     & (70, 0, 0) \\
orange      & (60, 35, 50)   & gold       & (65, 5, 35) \\
yellow      & (75, 5, 70)    & ---        & --- \\
\bottomrule
\end{tabular}
}
\label{tab:lab_prototypes}
\end{table}

\begin{table*}[t]
\centering
\caption{Performance comparison on MultiRefer. ZT: Zero-shot, ST: Single-task, MT: Multi-task, D: Distractor. Best results are bolded.}
\label{tab:detailed comparision on Multirefer}
\large
\resizebox{\textwidth}{!}{%
\begin{tabular}{lcccccccc|cc}
\toprule
\textbf{Method} & \textbf{ZT w/o D F1} & \textbf{ZT w/ D F1} & \textbf{ST w/o D F1@0.25} & \textbf{ST w/o D F1@0.5} & \textbf{ST w/ D F1@0.25} & \textbf{ST w/ D F1@0.5} & \textbf{MT F1@0.25} & \textbf{MT F1@0.5} & \textbf{Overall F1@0.25} & \textbf{Overall F1@0.5}\\ 
\midrule
3DVG-Trans+~\cite{Zhao20213DVGTransformerRM} & $87.1$ & $45.8$ & – & $27.5$ & – & $16.7$ & – & $26.5$- & - & $25.5$ \\
D3Net (Grounding)~\cite{Chen2021D3NetAS} & $81.6$ & $32.5$ & – & $36.6$ & – & $23.3$ & – & $35.0$ & - & $32.2$ \\
3DJCG (Grounding)~\cite{Cai20223DJCGAU} & $\mathbf{94.1}$ & $66.9$ & – & $26.0$ & – & $16.7$ & – & $26.2$ & - & $26.6$ \\
M3DRef-CLIP~\cite{zhang2023multi3drefer} & $81.8$ & $39.4$ & $53.5$ & $47.8$ & $34.6$ & $30.6$ & $43.6$ & $37.9$ & $42.8$ & $38.4$ \\
\midrule
Chat-Scene~\cite{huang2024chat} & $90.3$ & $62.6$ & $82.9$ & $75.9$ & $49.1$ & $44.5$ & $45.7$ & $41.1$ & $57.1$ & $52.4$ \\ 
Ours & $92.6$ & $69.3$ & $82.1$ & $75.6$ & $\mathbf{52.6}$ & $\mathbf{48.4}$ & $\mathbf{50.0}$ & $\mathbf{46.0}$ & $\mathbf{60.0}$ & $\mathbf{55.8}$ \\
\bottomrule
\end{tabular}%
}
\end{table*}

\begin{table}[t]
\centering
\caption{Performance comparison on the ScanRefer.}
\label{tab:more detailed results on Scanrefer}
\resizebox{0.85\linewidth}{!}{
\begin{tabular}{lcccccc}
\toprule
\multirow{2}{*}{\textbf{Method}} &
\multicolumn{2}{c}{\textbf{Unique}} & 
\multicolumn{2}{c}{\textbf{Multiple}} & 
\multicolumn{2}{c}{\textbf{Overall}} \\
\cmidrule(lr){2-3} \cmidrule(lr){4-5} \cmidrule(lr){6-7}
 & Acc@0.25 & Acc@0.5 & Acc@0.25 & Acc@0.5 & Acc@0.25 & Acc@0.5 \\
\midrule
Chat-scene~\cite{huang2024chat} & $\mathbf{89.6}$ & $82.5$ & $47.8$ & $42.9$ & $55.5$ & $50.2$ \\
Ours & $88.6$ & $81.9$ & $\mathbf{50.8}$ & $\mathbf{45.8}$ & $\mathbf{57.8}$ & $\mathbf{52.3}$ \\
\bottomrule
\end{tabular}}
\end{table}

\section{More Details about Experiment}
\label{experiment in appendix}

\subsection{Implementation Details}
\label{more implementation details}

In Round 1, we fine-tune the baseline model using the original training data plus ~10k counterfactual scenes and ~60k QA pairs.
In Round 2, we use the Round 1 model to identify new failure cases, generate an additional ~3k counterfactual scenes and ~18k QA pairs, and further fine-tune the model to obtain the Round 2 model.
For all fine-tuning stages, we train for $5$ epochs using the AdamW optimizer. The learning rate is $3e^{-5}$, with a weight decay of $0.01$ and a global batch size of $32$. All experiments are conducted on 4 NVIDIA H100 (80 GB) GPUs.
Our training follows an iterative refinement paradigm. In Round 1, the model is fine-tuned on the initial batch of DEER-3D counterfactual scenes. The resulting model is then used to automatically identify new failure cases on the training set; these cases are subsequently edited to produce an expanded set of counterfactual samples for Round 2. This second-stage training further corrects the model’s residual grounding errors.
To improve stability during optimization, especially given the distribution shift introduced by edited scenes, we apply an Exponential Moving Average (EMA) over model parameters when training the final Round 2 model. EMA smoothing helps regularize updates, reduces variance in later training stages, and yields a more robust final checkpoint with improved grounding accuracy.

\subsection{Computation Overhead}
 On 4 NVIDIA H100 GPUs, Round 1 introduces only ~3\% additional overhead (+4 min/epoch; 10.1h vs. 9.8h baseline). Round 2 adds around 3-4 more hours since it further fine-tunes on a smaller set of newly identified failure cases, while Round 1 alone already captures the majority of the gain (+4.4/+1.3 across two rounds on ScanRefer Acc@0.50). Regarding the comparison with random augmentation: under equal data scale, it has roughly the same per-epoch cost as DEER-3D but yields lower performance (50.9\% vs. 52.3\% Acc@0.50).

\begin{figure}[t]
    \centering
\includegraphics[width=0.65\linewidth]{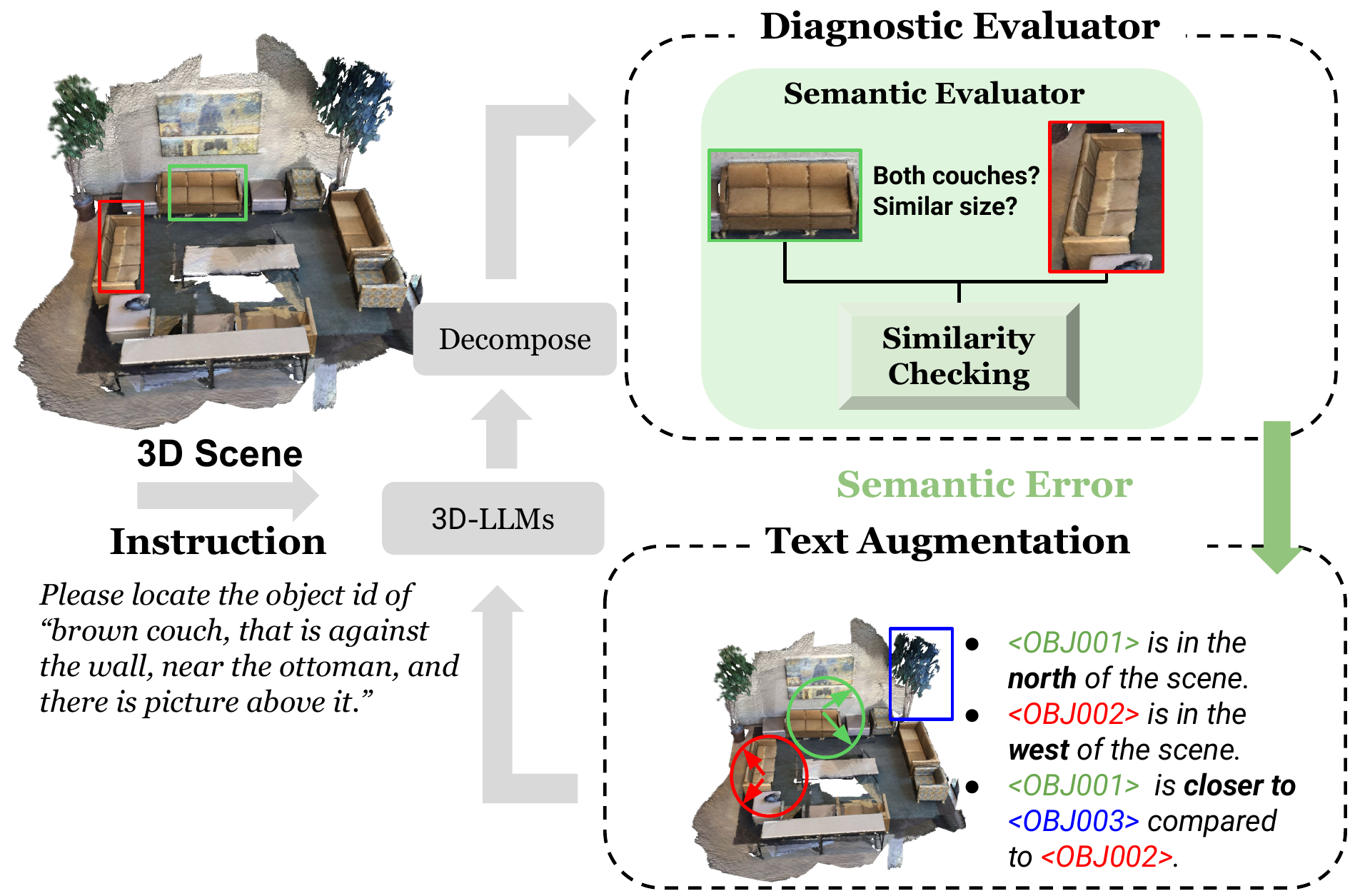}
    \caption{Semantic error detection followed by targeted text augmentation to disambiguate similar objects.}
    \label{fig:text_aug_architecture}
\end{figure}

\subsection{Performance on Scene Understanding Tasks}
\label{suppl for scene understanding}
Beyond 3D grounding, we further assess our models on a suite of general scene understanding tasks, including Scan2Cap~\cite{chen2021scan2cap}, ScanQA~\cite{azuma2022scanqa}, and SQA3D~\cite{ma2022sqa3d}.
For Scan2Cap, we evaluate captioning quality using CIDEr@0.5 (C@0.5) and BLEU-4@0.5 (B-4@0.5), following the standard practice in 3D captioning benchmarks.
For ScanQA, we adopt the official evaluation metrics of CIDEr (C)~\cite{vedantam2015cider} and BLEU-4 (B-4)~\cite{papineni2002bleu} to measure answer relevance and linguistic fidelity.
Finally, for SQA3D, we report Exact Match accuracy (EM) and its refined variant EM-R~\cite{huang2023embodied}, which provides a more granular assessment of spatially grounded question answering. Our method achieves consistent improvements over the baseline across all benchmarks, indicating that the benefits of our approach extend beyond grounding-specific settings.

\begin{table}[t]
\centering

\begin{minipage}[t]{0.48\linewidth}
\centering
\caption{Performance comparison on 3D grounding benchmarks.}
\label{tab:text_aug results}
\resizebox{\linewidth}{!}{
\begin{tabular}{lcc}
\toprule
\multirow{2}{*}{\textbf{Method}} &
\multicolumn{2}{c}{\textbf{ScanRefer}} \\
\cmidrule(lr){2-3}
 & Acc@0.25 & Acc@0.5 \\
\rowcolor{gray!10}Chat-Scene~\cite{huang2024chat} 
& 55.5 & 50.2 \\
Random-Text & 55.8 & 50.3 \\
DEER-3D+Text-Aug
& \textbf{56.4}
& \textbf{50.8} \\
\bottomrule
\end{tabular}}
\end{minipage}
\hfill
\begin{minipage}[t]{0.45\linewidth}
\centering
\caption{Iterative refinement beyond Round 2 shows saturated performance.}
\label{tab:round3_appendix}
\resizebox{\linewidth}{!}{
\begin{tabular}{lcc}
\toprule
\multirow{2}{*}{\textbf{Method}} &
\multicolumn{2}{c}{\textbf{ScanRefer}} \\
\cmidrule(lr){2-3}
 & Acc@0.25 & Acc@0.5 \\
\midrule
Ours (Round 1) & 57.8 & 55.8 \\
Ours (Round 2) & \textbf{58.6} & \textbf{56.8} \\
Ours (Round 3) & 58.6 & 56.7 \\
\bottomrule
\end{tabular}}
\end{minipage}

\end{table}

\subsection{More Results on Grounding Task}
\label{More Results on Grounding Task}
In the main paper, we report only the overall grounding performance on ScanRefer and Multi3DRefer. For completeness, we provide per–subset results for these benchmarks in~\cref{tab:detailed comparision on Multirefer} and~\cref{tab:more detailed results on Scanrefer}, respectively.
In particular, we observe that the improvements of DEER-3D are particularly obvious in the Multiple subset of ScanRefer and multi-task in MultiRefer, where multiple objects of the same category appear in the scene. 
This setting is more challenging because models must rely on fine-grained spatial relations or attribute cues to disambiguate between similar objects, rather than exploiting dataset-level biases. The results further support our motivation that targeted counterfactual supervision is particularly beneficial in scenarios where grounding requires resolving object ambiguity.

To further investigate whether iterative refinement continues to improve performance beyond Round~2, we perform an additional round of error mining and fine-tuning using the Round~2 model as the starting point.
As shown in~\cref{tab:round3_appendix}, the resulting Round~3 model exhibits almost no improvement over Round~2 (differences within $\pm0.1$ across metrics). This suggests that the model has already corrected the major grounding failures by the second iteration, and that additional refinement yields diminishing returns. 
These results confirm our claim in the main paper that Round~2 represents a practical stopping point that achieves the best balance between accuracy gains and computational cost.

\subsection{More Analysis}
\label{more ablations}
\begin{figure*}[t]
    \centering    \includegraphics[width=\linewidth]{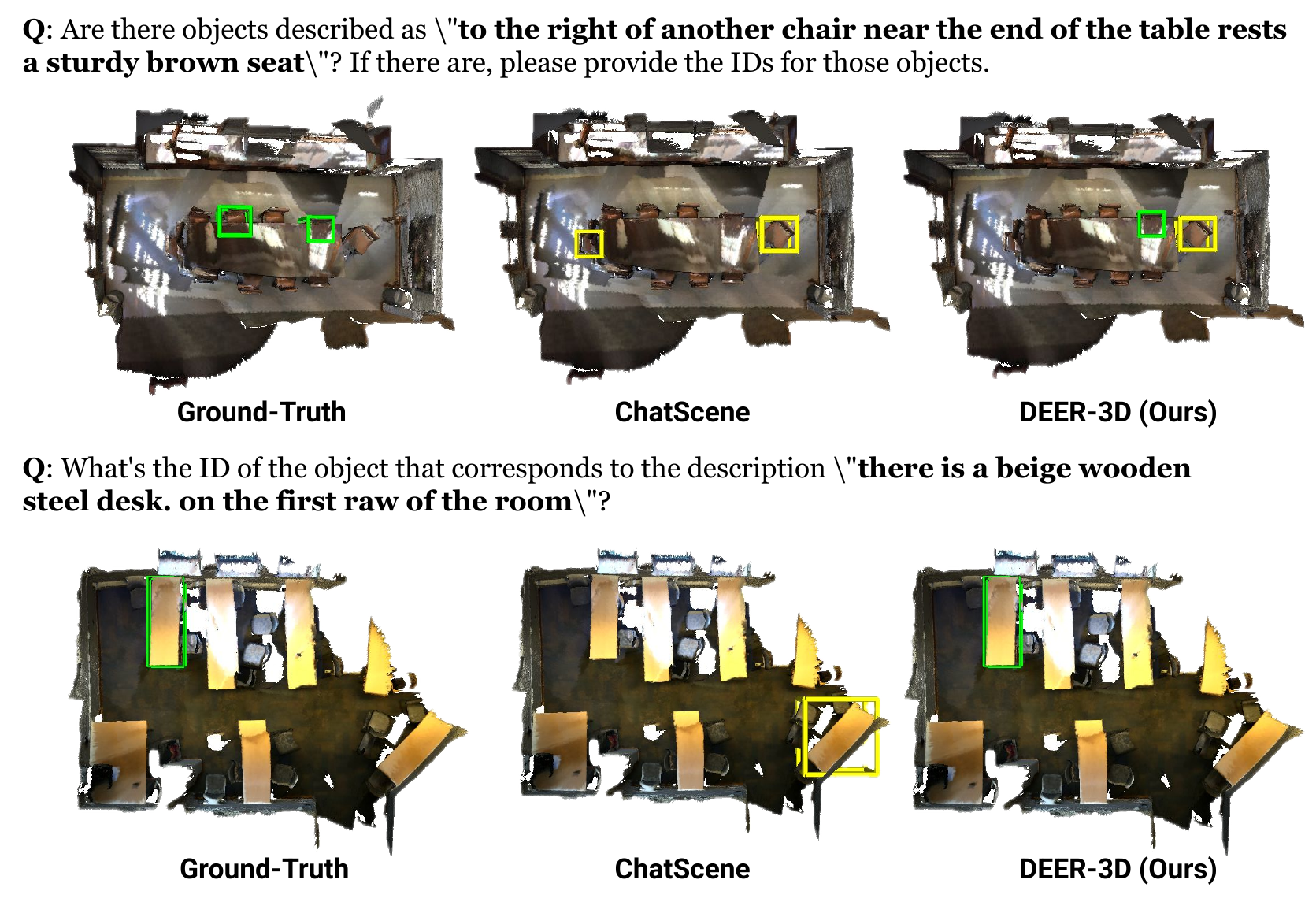}
      \textcolor{gtgreen}{\tikz{\draw[line width=0.8pt, color=gtgreen] (0,0) rectangle (0.25,0.25);}} ground-truth object;\,
\tikz{\draw[line width=0.8pt, color=yellow] (0,0) rectangle (0.25,0.25);} incorrectly predicted object
    \caption{Qualitative examples showcasing DEER-3D's improved grounding ability.}
\label{fig:qualitative_appendix}
\end{figure*}

\cref{fig:qualitative_appendix} illustrates that although \textsc{DEER}-3D is explicitly trained on specific geometric predicates (e.g., distance, rotation, and vertical relationships), its improved spatial sensitivity generalizes well to a broader range of spatial reasoning tasks.
In the top-row example, the query requires identifying a chair located "to the right of another chair." The baseline incorrectly identifies both chair positions due to a strong saliency bias toward the ends of the table. In contrast, our rotation- and distance-based augmentations encourage the model to encode accurate object poses and relative positions.
In the bottom-row example, given a query targeting a desk in the "first row," our distance-based counterfactual training effectively teaches the model to leverage precise 3D coordinate values rather than ignoring them. As a result, \textsc{DEER}-3D accurately interprets positional constraints, whereas the baseline simply predicts a visually prominent object of the correct category.

\subsection{Error-Driven Framework Evaluation}
\label{Error-Driven Framework Evaluation}
Beyond generating targeted 3D scene edits, our error-driven framework can also produce targeted text augmentations that disambiguate instructions. As shown in~\cref{fig:text_aug_architecture}, we design an experiment focusing on a subset of grounding failures where the predicted and ground-truth objects belong to the same semantic class and exhibit similar sizes and appearances. These failures typically arise from instructional ambiguity, where the textual description does not provide sufficient cues for the model to identify the correct instance.

Our diagnostic evaluator detects such cases by checking semantic consistency between the predicted and ground-truth objects (e.g., class match and size similarity). When an error is attributed to ambiguous language rather than visual misinterpretation, the framework triggers the text augmentation module instead of applying a 3D edit. This module generates precise disambiguating descriptions by incorporating contextual object references that uniquely characterize the target, such as: (1) global position cues (e.g., “located in the west of the scene”), (2) local relational attributes (e.g., “closer to $<$OBJ003$>$ than $<$OBJ002$>$”).

These augmentations enrich the instruction with predicate-level distinctions, enabling the model to differentiate between visually similar candidates without modifying the underlying scene. As shown in ~\cref{tab:text_aug results}, simple random text perturbations yield minimal improvement, indicating that generic augmentation is insufficient for resolving ambiguity. In contrast, our targeted text augmentation provides better performance, demonstrating that error-driven, semantically grounded clarifications improve disambiguation. However, the overall gains remain relatively modest compared to our full editing pipeline, suggesting that text-only augmentation is helpful but inherently limited for 3D grounding tasks.







\section{Limitation and Future Work}
\label{Limitation and Future Work}
Despite the improvements introduced by DEER-3D, several broader challenges in current 3D-LLMs remain. Many grounding failures originate from limitations inherent to point-cloud–based 3D perception pipelines, such as incomplete geometry or reconstruction noise. The capability of the base 3D-LLM also fundamentally constrains the upper bound of achievable performance, as weaker reasoning or recognition ability limits the effectiveness of error diagnosis and correction. In addition, our current framework operates on datasets of relatively modest scale and primarily focuses on indoor environments, which restricts the magnitude and generality of performance gains. Finally, although our counterfactual edits introduce targeted and meaningful variations, they still cannot cover the long-tail complexity of real-world 3D scenes involving deformable objects, human interactions, or dynamic environments. Future work may explore scaling DEER-3D to larger and more diverse 3D corpora, leveraging stronger 3D-LLMs for better perception and reasoning, and enriching the editing module to support more complex and varied scenes.